\documentclass[journal]{IEEEtran}

\usepackage[utf8]{inputenc}
\usepackage[margin=0.70in]{geometry}
\usepackage{mathtools}
\usepackage{amssymb}
\usepackage{algorithm}
\usepackage{algpseudocode}
\usepackage{comment}
\usepackage{lipsum}
\usepackage{graphicx}
\usepackage{subcaption}
\usepackage{multicol}
\usepackage{xcolor}
\usepackage{amsthm}
\usepackage{Symbol}
\usepackage{algcompatible}
\usepackage{siunitx}
\usepackage{booktabs}
\usepackage{color, colortbl}
\usepackage{enumitem}
\usepackage{cuted}

\usepackage[noadjust]{cite}

\makeatletter
\newcommand*{\rom}[1]{\expandafter\@slowromancap\romannumeral #1@}
\newtheorem{proposition}{Proposition}
\newtheorem{theorem}{Theorem}
\newtheorem{remark}{Remark}

\newtheorem{assumption}{Assumption}
\newtheorem{corollary}{Corollary}
\newtheorem{lemma}{Lemma}

\def\ccal{\mathcal}
\def\bf{\mathbf}
\def\bb{\mathbb}

\begin{document}
\title{Online Graph Filtering Over Expanding Graphs}
\author{Bishwadeep~Das,~\IEEEmembership{Student~Member,~IEEE}
        and~Elvin~Isufi,~\IEEEmembership{Senior Member,~IEEE}
        \thanks{Preliminary results have appeared in \cite{das2022online}. This research is supported by the TTW-OTP project GraSPA (project number 19497) financed by the Dutch Research Council (NWO) and by the TU Delft AI programme. The authors are with the Faculty of Electrical Engineering, Mathematics and Computer Science, Delft University of Technology, The Netherlands. e-mails: \{b.das,e.isufi-1\}@tudelft.nl}}
\maketitle

\begin{abstract}
\textbf{Graph filters are a staple tool for processing signals over graphs in a multitude of downstream tasks. However, they are commonly designed for graphs with a fixed number of nodes, despite real-world networks typically grow over time. This topological evolution is often known up to a stochastic model, thus, making conventional graph filters ill-equipped to withstand such topological changes, their uncertainty, as well as the dynamic nature of the incoming data. To tackle these issues, we propose an online graph filtering framework by relying on online learning principles. We design filters for scenarios where the topology is both known and unknown, including a learner adaptive to such evolution. We conduct a regret analysis to highlight the role played by the different components such as the online algorithm, the filter order, and the growing graph model. Numerical experiments with synthetic and real data corroborate the proposed approach for graph signal inference tasks and show a competitive performance w.r.t. baselines and state-of-the-art alternatives.} 
\end{abstract}

\begin{IEEEkeywords}
Graph filters, graph signal processing, online learning.
\end{IEEEkeywords}

\IEEEpeerreviewmaketitle
\section{Introduction}

Graph filters are a well-established tool to process network data and have found use in a variety of applications, including node classification \cite{dong2020graph,berberidis2018adaptive}, signal interpolation \cite{sandryhaila2013discrete}, and product recommendation \cite{isufi2021accuracy}. They are a flexible parametric and localized operator that can process signals defined over the nodes through a weighted combination of successive shifts between neighbours \cite{isufi2024graph}. Being the analogue of filters in discrete signal processing, graph filters can be interpreted in the graph frequency domain \cite{sandryhaila2014discrete}. Compared to other tools such as graph kernels \cite{romero2016kernel}, filters do not need any prior knowledge about the data w.r.t. the topology when used for inference tasks.
\par Most of the filters in literature are designed out over graphs with a fixed number of nodes \cite{isufi2024graph} despite graphs often growing through the addition of nodes, sometimes sequentially over time \cite{barabasi_emergence_1999,barabasi2016network}. An example is collaborative filtering in recommender systems where new users continuously join an existing network, e.g., a social network recommendation \cite{wang2010graph} or an abstract user-user collaborative filter network \cite{huang_rating_2018}. Such an expanding graph setting poses a three-fold challenge: $(i)$ The data comes in a streaming nature, i.e., we do not have access to all the incoming nodes at once. This requires an on-the-fly filter design as batch-based solutions are no longer an alternative. $(ii)$ The topology may evolve slowly or rapidly; hence, influencing the online filter design.  $(iii)$ The data over the incoming nodes is not guaranteed to follow a well-known distribution, thus requiring an adaptation of the filter to the task at hand. Often times, we may not even know how the incoming nodes connect to the existing graph. Typically, this happens in the absence of information for the incoming node, i.e., in pure cold-start recommendation, where we know nothing about user preferences, but we need to recommend items nevertheless \cite{silva2019pure}. The users may consume items later on, which can be used to infer their attachment but this can take time. Such challenges limit existing graph data processing methods which rely on the knowledge of the topology \cite{ortega_graph_2018}. Another example where these challenges occur is in epidemic spreading over networks. We want to predict the future number of active cases for a city that is not yet affected but anticipates some cases shortly after. It may be difficult to obtain the underlying connections that influence the disease spread; hence, using statistical models is typically an option \cite{erdos_evolution_1961,barabasi_emergence_1999}. In this scenario, filter design should account for the evolving topological model as well as for the data over it. This is possible by building upon online learning principles where the learning models are updated based on the incoming data stream \cite{orabona2019modern,hazan2016introduction}. 
 \par Existing works dealing with online learning over expanding graphs can be divided into Attachment, Feature Aware Methods, and Stochastic Methods. \textit{Attachment and feature aware methods} know the connectivity of the incoming nodes and their features. For example, the work in \cite{shen_online_2019} performs online node regression over fixed-size graphs by using their connectivity pattern to generate random kernel features \cite{rahimi2007random}. An extension of this is the work in \cite{zong2021online} which considers multi-hop connectivity patterns. Works like \cite{money2021online}, \cite{money2023sparse}, and \cite{shafipour2020online} track changing attachment patterns over time but for graphs with a fixed number of node, which can be relevant for a large-scale setting. Another instance of online processing on expanding graphs is the work in \cite{venkitaraman_recursive_2020} which obtains embeddings for signals over expanding graphs. Some works such as \cite{chen2014semi} classify an incoming node by using its features and the filter trained over the existing graph.  Then, there are works such as \cite{dornaika2017efficient} that classify a stream of incoming nodes by using their features to estimate the attachment. The work in \cite{jian2018toward} classifies incoming nodes using spectral embeddings updated from the known attachment information. The kernel-based methods in this category \cite{shen_online_2019,zong2021online} rely on pre-selecting a suitable kernel that can fit the data which may be challenging to obtain. Additionally, the works in \cite{nassif2017graph,hua2020online,nassif2018distributed,elias2020adaptive} develop distributed solutions to estimate the filter parameters locally at each node. Differently, in this paper, we work with a centralized approach to estimate the filter, as we focus more on the expanding graph scenario. All in all, these methods concern either a graph with a fixed or a streaming number of nodes but with available attachment or feature information that may be unavailable.
 
\textit{Stochastic methods} deal with unknown incoming node attachment and use models for it. For example, \cite{dasfiltering2020} uses heuristic stochastic attachment model to design graph filters only for one incoming node, while \cite{liu2018streaming} learns an embedding by using a stochastic attachment to influence the propagation. In our earlier works \cite{das2022learning}, \cite{das2022task}, we learn the attachment behaviour for inference with a fixed filter. However, this approach is limited to studying the effect of one node attaching with unknown connectivity and it assumes a pre-trained filter over the existing graph. Differently, here we consider the filter design over a stream of incoming nodes.
\par We perform online graph filtering over a stream of incoming nodes when the topology is both known and unknown. Our contribution is threefold:
\begin{enumerate}
\item We develop an online filter design framework for inference over expanding graphs. This is done by casting the inference problem as a time-varying loss function over the existing topology, data, and the incoming node attachment. Subsequently, we update the filter parameters via online learning principles. 
\item We adapt the online filter design problem to two scenarios: $(i)$ the \textit{deterministic} setting where the connectivity of each incoming node is available; $(ii)$ the \textit{stochastic} setting where this connectivity is unavailable. For both settings, we conduct a regret analysis to discuss the influence of the incoming node attachment and the role of the graph filter.
\item We develop an online ensemble and adaptive stochastic update where, in addition to the filter parameters, we also learn the combination parameters of the different stochastic attachment rules. This concerns the stochastic setting where a single attachment model might be insufficient. We also discuss the regret in this setting and analyze how the ensemble affects it.
\end{enumerate}
\par We corroborate the proposed approach with numerical experiments on synthetic and real data from recommender systems and COVID cases prediction. Results show that the online filters perform better than other alternatives like kernels or pre-trained filters; and, that stochastic online filters can also perform well w.r.t. deterministic approaches.
\par This paper is structured as follows. Sec. \ref{Section PF} elaborates on the sequentially expanding graph scenario, along with the basic formulation of online inference with graph filters. Sec. \ref{Section Deterministic} and \ref{Section Stochastic} contain the online learning methods and their respective analysis in the deterministic and in the stochastic setting, respectively. Sec. \ref{Section Results} contains the numerical results, while Sec. \ref{Section Conclusion} concludes the paper. All proofs are collected in the appendix.

\section{Problem Formulation}\label{Section PF}
Consider a starting graph $\ccalG_0=\{\ccalV_0,\ccalE_0\}$ with node set $\ccalV_0=\{v_{0,1},\ldots,v_{0,N_0}\}$ of $N_0$ nodes, edge set $\ccalE_0$, and adjacency matrix $\bbA_0\in\reals^{N_0\times N_0}$, which can be symmetric or not, depending on the type of graph (undirected or directed). Let $v_1,\ldots,v_T$ be a set of $T$ sequentially incoming nodes where at time $t$, node $v_t$ attaches to graph $\ccalG_{t-1}$ forming the graph $\ccalG_t=\{\ccalV_t,\ccalE_t\}$ with $N_t=N_0+t$ nodes, $M_t$ edges, and adjacency matrix $\bbA_t\in\reals^{N_t\times N_t}$. The connectivity of $v_{t}$ is represented by the attachment vector $\bba_{t}=[a_1,\ldots,a_{N_{t-1}}]^{\top}\in\reals^{N_{t-1}}$, where a non-zero element implies a directed edge from $v\in\ccalV_{t-1}$ to $v_t$. This connectivity suits inference tasks at $v_{t}$, where the existing nodes influence the incoming ones. This is the case of cold-starters in graph-based collaborative filtering \cite{huang_rating_2018,isufi2021accuracy}. Here, the nodes represent existing users, the edges capture similarities among them (e.g., Pearson correlation), and a cold starter is a new node that attaches to this user-user graph. The task is to collaboratively infer the preference of the cold-starter from the existing users \cite{schein2002methods}.

Depending on the availability of $\bba_t$, we can have a deterministic attachment setting or a stochastic attachment setting. In a deterministic setting, the incoming node attachment vector $\bba_t$ is known or it is estimated when $v_t$ appears. This occurs in growing physical networks or in collaborative filtering where side information is used to establish the connectivity \cite{huang_rating_2018}. The expanded adjacency matrix $\bbA_{t}\in\reals^{N_{t}\times N_{t}}$ reads as 
    \begin{equation}\label{topology update}
        \bbA_{t}=\begin{bmatrix}
        \bbA_{t-1} & \mathbf0_{N_{t-1}} \\
        \bba_{t}^{\top} & 0 \\\end{bmatrix}
    \end{equation}where $\bbA_{t-1}$ is the $N_{t-1}\!\times\!N_{t-1}$ adjacency matrix and $\mathbf0_{N_{t-1}}$ is the all-zero vector of size $(N_{t-1})$. In a stochastic setting, $\bba_{t}$ is unknown (at least before inference), which is the typical case in cold start collaborative filtering \cite{das2022task}. A new user/item enters the system and we have neither side information nor available ratings to estimate the connectivity. The attachment of $v_{t}$ is modelled via stochastic models from network science \cite{barabasi_emergence_1999}. Node $v_{t}$ attaches to $v_i\in\ccalV_{t-1}$  with probability $p_{i,t}$ forming an edge with weight $w_{i,t}$. The probability vector $\bbp_t=[p_{1,t},\ldots,p_{N_{t-1},t}]^{\top}\in\reals^{N_{t-1}}$ and the weight vector $\bbw_t=[w_{1,t},\ldots,w_{N_{t-1},t}]^{\top}\in\reals^{N_{t-1}}$ characterize the attachment and imply that $[\bba_{t}]_i=w_{i,t}$ with probability $p_{i,t}$, and zero otherwise. We consider vector $\bba_{t}$ be composed of independent, weighted Bernoulli random variables with respective mean and covariance matrix
\begin{equation}\label{eq.attachment_rule}
        \E{\bba_{t}}=\bbp_t\circ\bbw_t~;~
\bbSigma_{t}=\diag(\bbw_t^{\circ2}\circ\bbp_t\circ(\mathbf 1-\bbp_t))
    \end{equation}
where $\text{diag}(\bbx)$ is a diagonal matrix with $\bbx$ comprising the diagonal elements, and  $\bbx^{\circ2}=\bbx\circ\bbx$ is the element-wise product of a vector $\bbx$ with itself. The new adjacency matrix for a realization $\bba_t$ is the same as in \eqref{topology update}. The attachment is revealed after the inference task; e.g., after a cold start user has consumed one or more items and we can estimate it.

\subsection{Filtering over Expanding Graphs}
Let $\bbx_t\in\reals^{N_t}$ be the graph signal over graph $\ccalG_t$, which writes in terms of the previous signal $\bbx_{t-1}\in\reals^{N_{t-1}}$ as $\bbx_{t}=[\bbx_{t-1} ,x_t]^{\top}$ with $x_t$ being the signal at the latest incoming node $v_t$. To infer $x_t$, we consider the temporary graph signal $\tilde{\bbx}_t=[\bbx_{t-1},0]^{\top}$ where the zero at $v_t$ indicates that its value is unknown. To process such signals we use graph convolutional filters, which are linear and flexible tools for processing them \cite{isufi2024graph}. A filter of order $K$ acts on $ \tilde{\bbx}_t$ to generate the output $\tilde{\bby}_t$ on graph $\ccalG_t$ as
    \begin{equation}\label{FIRop}
 \tilde{\bby}_t=\sum_{k=0}^{K}h_{k}\bbA_{t}^k\tilde{\bbx}_t
    \end{equation}where $h_{k}$ is the weight given to the $k$th shift $\bbA_{t}^k\tilde{\bbx}_t$. Substituting the $k$th adjacency matrix power
    \begin{equation}
        \bbA_{t}^k=\begin{bmatrix}
        \bbA_{t-1}^{k} & \mathbf0_{N_{t-1}} \\
        \bba_{t}^{\top}\bbA_t^{k-1} & 0 \\\end{bmatrix}
    \end{equation}into \eqref{FIRop}, we write the filter output as
    \begin{equation}\label{filter output}
        \tilde{\bby}_t=\begin{bmatrix}
        \sum_{k=0}^Kh_{k}\bbA_{t-1}^{k}{\bbx}_t \\
        \bba_{t}^{\top}\sum_{k=1}^Kh_{k}\bbA_{t-1}^{k-1}{\bbx}_t\\\end{bmatrix}
    \end{equation}where we grouped w.l.o.g. the output at the incoming node $v_t$ in the last entry. I.e.,
    \begin{equation}\label{eq. output at v_t}
     {[\tilde{\bby}_t]}_{N_t}:=\hat x_t=\bba_{t}^{\top}\sum_{k=1}^K h_{k}\bbA_{t-1}^{k-1}{\bbx}_t=\bba_{t}^{\top}\bbA_{x,t-1}\bbh.   
    \end{equation}Here $\bbh=[h_1\ldots,h_K]^{\top}\in\reals^{K}$ collects the filter coefficients and $\bbA_{x,t-1}=[{\bbx}_{t},\bbA_{t-1}{\bbx}_{t},\ldots,\bbA_{t-1}^{K-1}{\bbx}_{t}]\in\reals^{N_{t-1}\times K}$ contains the higher-order shifts of ${\bbx}_{t}$. The coefficient $h_0$ does not play a role in the output $\hat x_t$, thus the zero in the $N_t$th position of $\tilde{\bbx}_t$ does not influence the inference task on the incoming node. In the stochastic setting, the output is random as it depends on the attachment rule. In turn, this needs a statistical approach to characterize both the filter and its output. We shall detail this in Section \ref{Section Stochastic}.
\begin{figure}[t]
{\centering
\includegraphics[trim=0 0 0 0,clip,width=0.5\textwidth]{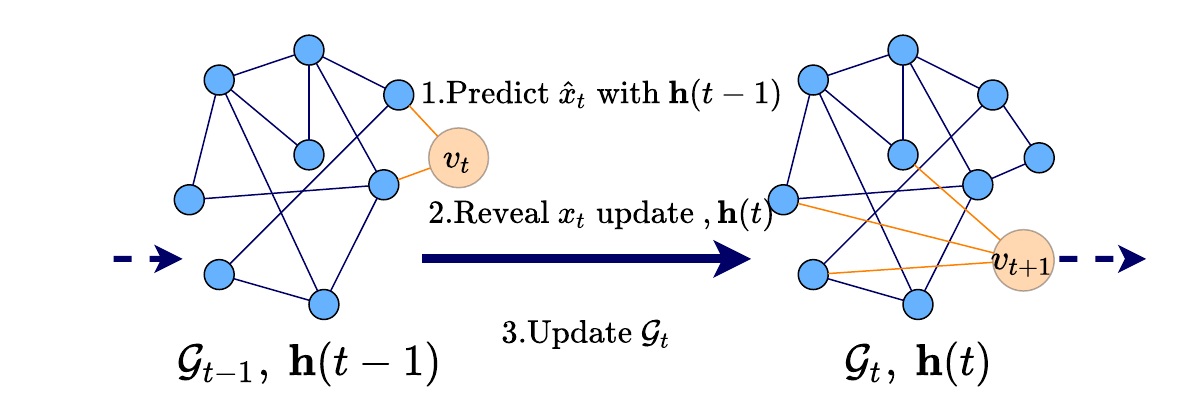}%
\caption{\small{Online filter learning process at time $t$ through the addition of node $v_t$ with signal $x_t$ and the update of the filter $\bbh(t)$. (Left) node $v_t$ attaches to the previous graph $\ccalG_{t-1}$ forming $\ccalG_t$; the edges in blue denote the existing edges while those in orange denote the edges formed by the incoming node; (Centre)  Signal $x_t$ is predicted, then the true value $x_t$ is revealed, and the filter parameter $\bbh(t)$ is updated from $\bbh(t-1)$; (Right) the next node $v_{t+1}$ attaches to $\ccalG_{t}$.}}
\label{freq_plots_recsys}} 
\end{figure}

\begin{remark}
    The above discussion considers one incoming node at a time which is common in the streaming setting. The analysis can be extended to multiple nodes arriving at a certain time interval. Here, we consider inference tasks where the existing nodes affect the incoming streaming ones. For tasks where the influence is bidirectional, the adjacency matrix in \eqref{topology update} is symmetric and the analysis follows analogously. One way to do this is to build on \cite{dasfiltering2020}, where we discuss the case for a single incoming node.
\end{remark}
\subsection{Online Filter Learning}
Our goal is to process signal $\tilde{\bbx}_t$ to make inference on the incoming nodes $v_t$ by designing the filters in \eqref{FIRop}. We consider a data-driven setting where we estimate the filter parameters from a training set $\ccalT=\{v_t,x_t,\bba_t\}_{t=1:T}$ in which each datum comprises an incoming node $v_t$, its signal $x_t$, and the attachment vector $\bba_t$. Given set $\ccalT$, we find the filter parameters $\bbh$ by solving
\begin{align}
    \begin{split}
        \underset{\bbh\in\reals^{K}}{\text{argmin}}\sum_{t=1}^T f_{t}(\hat x_t,x_{t};\bbh)+r(\bbh)
    \end{split}
\end{align}
where $f_{t}(\hat x_t,x_{t};\bbh):=f_t(\bbh,x_t)$ measures the goodness of fit between the prediction $\hat x_t$ and the true signal $x_t$ and $r(\bbh)$ is a regularizer. For example, $f_t(\cdot,\cdot)$ can be the least-squares error for regression problems such as signal denoising or interpolation; or the logistic error for classification problems such as assigning a class label to node $v_t$. For convex and differentiable $f_t(\cdot,\cdot)$ and $r(\bbh)$, we can find an optimal filter that solves the batch problem over $\ccalT$. However such a solution is not ideal since the incoming nodes $v_t$ are streaming and evaluating a new batch for each $v_t$ is computationally demanding. A batch solution also suffers in non-stationary environments where the test set distribution differs from $\ccalT$.
Targeting a non-stationary setting with incoming nodes, we turn to online learning to update the filter parameters on-the-fly \cite{hazan2016introduction}. 
\par We initialize the filter before the arrival of incoming nodes, $\bbh(0)$ by training a filter over $\ccalG_0$, using $\bbA_0$ and $\bbx_0$. The training follows a regularized least square problem with $\ell_2$ norm squared loss on the filter. We call this \emph{Pre-training}. In high-level terms, the online filter update proceeds at time $t$ as follows:
\begin{enumerate}
\item  The environment reveals the node $v_t$ and its attachment $\bba_t$ in the deterministic setting.
\item We use the filter at time $t$, $\bbh(t-1)$ to infer the signal value $\hat x_t$ at the incoming node using \eqref{eq. output at v_t}.
\item The environment reveals the loss as a function of the filter $\bbh(t-1)$ and the true signal $x_t$ as
\begin{equation}
l_t(\bbh,x_t)=f_t(\bbh,x_t)+ r(\bbh)
\end{equation}which is evaluated at $\bbh(t-1)$. 
\item We update the filter parameters $\bbh(t)$ based on the loss and the current estimate $\bbh(t-1)$.
\item In the stochastic setting, the true attachment $\bba_t$ is revealed.
\end{enumerate}
With this in place, our problem statement reads as follows:
\par\smallskip\noindent\textbf{Problem statement}: \textit{Given the starting graph $\ccalG_0=\{\ccalV_0,\ccalE_0\}$, adjacency matrix $\bbA_0$, graph signal $\bbx_0$, and the training set $\ccalT$, our goal is to predict online a sequence of graph filters $\{\bbh(t)\}$ w.r.t. loss functions $l_t(\bbh,x_t)$ to process signals at the incoming nodes for both the deterministic and the stochastic attachments.}
 \begin{algorithm}[!t]
	\caption{Deterministic Online Graph Filtering (\textbf{D-OGF})}
	\begin{algorithmic}
	\STATE \textbf{Input:} Graph $\ccalG_0$, $\bbA_0$, $\bbx_0$, $\ccalT=\{v_t,x_t,\bba_t\}_{t=1:T}$.
	\STATE \textbf{Initialize:} Pre-train $\bbh(0)$ over $\ccalG_0$ using $\bbA_0$, $\bbx_0$.
	\FOR{$t=1:T$}
	\STATE Obtain $v_t$ and true connection $\bba_{t}$, update $\bbA_t$
	\State Predict $ \hat x_t=\bba_{t}^{\top}\bbA_{x,t-1}\bbh(t-1)$ ~(cf. \eqref{eq. output at v_t})
	\STATE Reveal loss $l_t(\bbh,x_t)$~(cf. \eqref{instant loss})
	\STATE Update $\bbh(t)$ using \eqref{online update}
	\STATE Update $\bbx_{t}$ 
	\ENDFOR
	\end{algorithmic} 
	\label{Online deterministic}
\end{algorithm}

\section{Deterministic Online Filtering}\label{Section Deterministic}
Targeting regression tasks\footnote{For classification tasks, we can consider the surrogate of the gradient of the logistic loss which is also convex and differentiable, as seen in \cite{vlaski2023networked}.}, we can take $f_t(\bbh,x_t)$ as the squared error and $r(\bbh)$ as the scaled $l_2$-norm penalty to define the loss
\begin{equation}\label{instant loss}
l_t(\bbh,x_{t})=\frac{1}{2}(\bba_{t}^{\top}\bbA_{x,t-1}\bbh-x_{t})^2+\mu||\bbh||_2^2
\end{equation}where $\mu>0$. 
For the online update, we perform projected online gradient descent \cite{orabona2019modern}, which comprises one projected gradient descent step evaluated at $\bbh(t-1)$ as 
\begin{equation}\label{online update}
    \bbh(t)=\underset{\ccalH}{\Pi}(\bbh(t-1)-\eta\nabla_h l_t(\bbh,x_t)|_{\bbh(t-1)})
\end{equation}with set ${\ccalH}$ bounding the filter energy $\ccalE(\bbh)=||\bbh||_2^2$ and $\underset{\ccalH}{\Pi}(\cdot)$ denotes the projection operator on $\ccalH$. Here, $\eta>0$ is the step size, and the gradient has the expression
\begin{equation}\label{det gradient}
 \nabla_h l_t(\bbh,x_t)=(\bba_{t}^{\top}\bbA_{x,t-1}\bbh-x_{t})\bbA_{x,t-1}^{\top}\bba_{t}+2\mu\bbh.   
\end{equation}
The gradient depends on $\bba_t$ through the term $(\hat x_t-x_t)\bbA_{x,t-1}^{\top}\bba_{t}$. Operation $\bbA_{x,t-1}^{\top}\bba_{t}$ is a weighted combination of only those columns of $\bbA_{x,t-1}^{\top}$ where the corresponding entry of $\bba_t$ is non-zero. In turn, each column of $\bbA_{x,t-1}^{\top}$ contains shifted graph signals at each node, which get scaled by the difference between the predicted and the true signal $x_t$, ultimately, indicating that a larger residue leads to a larger gradient magnitude. The online learner in \eqref{online update} updates the filter parameters for every incoming node. 
\par Algorithm \ref{Online deterministic} summarizes the learning process. The computational complexity of the online update at time $t$ is of order $\ccalO(K(M_{t}+M_{\textnormal{max}}))$, where $M_{\textnormal{max}}$ is the maximum number of edges formed by $v_t$ across all $t$. Note that $M_{\textnormal{max}}\ll N_{t-1}$, i.e., the maximum number of edges formed by any incoming node is smaller than the existing number of nodes. Appendix \ref{AppB} breaks down this complexity.

\smallskip
\par \noindent\textbf{Regret analysis.} We analyze the deterministic online graph filtering algorithm to understand the effect of the filter updates and how the expanding graph influences it. Specifically, we conduct a regret analysis that quantifies the performance difference between the online updates and the static batch solution where all the incoming node information is available. The normalized regret w.r.t. a fixed filter $\bbh^{\star}$ is defined as
    \begin{equation}\label{regret basic}
        \frac{1}{T}R_{T}(\bbh^{\star})=\frac{1}{T}\sum_{t=1}^{T}l_t(\bbh(t-1),x_t)-l_t(\bbh^{\star},x_t)
    \end{equation}where $\sum_{t=1}^Tl_t(\bbh(t-1),x_t)$ is the cumulative loss incurred by the online algorithm. The regret measures how much better or worse the online algorithm performs over the sequence compared to a fixed learner. An upper bound on the regret indicates the worst-case performance and it is of theoretical interest. If this bound is sub-linear in time, the average regret tends to zero as the sample size grows to infinity, i.e., $\lim\limits_{T \to\infty}\frac{1}{T}R_{T}(\bbh^{\star})=0$  \cite{shalev2012online}. This indicates that the algorithm is learning. We assume the following.
\begin{assumption}\label{assumption1}
The incoming nodes form a maximum of $M_{max}\ll N_t$ edges for all $t$.
\end{assumption}
\begin{assumption}\label{assumption2}
 The attachment vectors $\bba_t$ and the stochastic model-based weight vectors $\bbw_t$ are upper-bounded by a scalar $w_h$. I.e., for all $t$ we have
 \begin{equation}
[\bba_t]_n \leq w_h,~ [\bbw_t]_n \leq w_h. 
 \end{equation}
\end{assumption}
\begin{assumption}\label{assumption3}
The filter parameters $\bbh$ are upper-bounded in their energy, i.e., $\ccalE(\bbh)=||\bbh||_2^2\leq H^2$.
\end{assumption}
\begin{assumption}\label{assumption4}
For all attachment vectors $\bba_t$, the residue $r_t=\bba_t^{\top}\bbA_{x,t-1}\bbh-x_{t}$ is upper-bounded. I.e., there exists a finite scalar $R > 0$ such that $|r_t|\leq R$. 
\end{assumption}
Assumption \ref{assumption1} holds for graphs in the real world. A node makes very few connections compared to the total number of nodes, i.e., the attachment vector $\bba_t$ will be sparse  with a maximum of $M_{max}$ non-zero entries. Note that our stochastic model does not take this into account. Assumption \ref{assumption2} bounds all edge-weights which is commonly observed. Assumption \ref{assumption3} ensures finite parameters which mean the filter output does not diverge. This can be guaranteed by projection on to $\ccalH$. Assumptions \ref{assumption3} and \ref{assumption4} imply bounded filter outputs. Then, we claim the following.
\begin{proposition}\label{Prop1}
\textit{Consider a sequence of Lipschitz losses  $\{l_t(\bbh,x_{t})\}$ with Lipschitz constants $L_d$, $[cf.~\eqref{instant loss}]$ and a learning rate $\eta$ $[cf. \eqref{online update}]$. Let also Assumptions \ref{assumption1}-\ref{assumption4} hold. The normalized static regret $R_{T}(\bbh^{\star})$ for the online algorithm generating filters $\{\bbh(t)\}\in\ccalH$ relative to the optimal filter $\bbh^{\star}\in\ccalH$ is upper-bounded as}
\begin{equation}\label{regret}
    \frac{1}{T}R_{T}(\bbh^{\star})\leq \frac{||\bbh^{\star}||_2^2}{2\eta T}+\frac{\eta}{2}L_d^2
\end{equation}with $L_d=RC+2\mu H$ where $||\bbA_{x,t-1}^{\top}\bba_t||_2\leq C$.
\end{proposition}
\par\noindent\textit{Proof}.~~From Lemma \ref{req} in Appendix \ref{AppB}, we have that the loss functions are Lipschitz. Then, we are in the setting of [Thm. 2.13,\cite{orabona2019modern}] from which the rest of the proof follows.\qed
\par\noindent There are two main filter-related factors that influence the regret bound in \eqref{regret}: the filter energy $H^2$  and the residual energy $R^2$. A smaller $H$ can lead to a lower bound but it can also increase the prediction error by constraining the parameter set too much. Moreover, a higher regularization weight $\mu$ also penalizes high filter energies $||\bbh||_2^2$. So, for the projected online learner with a high regularization weight $\mu$, a high $H$ can help the inference task, even if it increases the regret bound. Second, from Assumption~\ref{assumption4}, the residue $R$ is likely small when a filter approximates well the signal on the incoming node. This can happen when the signal values on the incoming node and the existing nodes are similar or when the existing topology and signals over it are expressive enough to represent the incoming node values. Examples of the latter are locally smooth graph signals that can be approximated by a low order filter $K$. For high values of $K$, all nodes have similar signals, implying that many potential attachment patterns can generate $x_t$. This would make that the manner of attachment irrelevant.

\begin{algorithm}[!t]
\caption{Stochastic Online Graph Filtering (\textbf{S-OGF})}
\begin{algorithmic} 
\STATE \textbf{Input:} Graph $\ccalG_0$, $\bbA_0$, $\bbx_0$, $\ccalT = \ccalT=\{v_t,x_t,\bba_t\}_{t=1:T}$
\STATE \textbf{Initialize:} Pre-train $\bbh^s(0)$ over $\ccalG_0$ using $\bbA_0$, $\bbx_0$.
\FOR{$t=1:T$}
\STATE Obtain $v_t$ and $\bbp_t$, $\bbw_t$ following preset heuristics
\STATE Predict $\hat x_{t}=(\bbw_t\circ\bbp_t)^{\top}\bbA_{x,t-1}\bbh^s(t-1)$
\STATE Incur loss $l_t^s(\bbh,x_t)$ [cf. \eqref{eq.stochLoss}]
\STATE Update $\bbh^s(t)$ using \eqref{stochastic update}
\STATE Reveal $\bba_t$, update $\bbA_t$ and $\bbx_t$
\ENDFOR
\end{algorithmic}
\label{stochastic algorithm}
\end{algorithm}

\section{Stochastic online filtering}\label{Section Stochastic}

Often, the true attachment for the incoming nodes is initially unknown and it is only revealed afterwards. This is the case with rating prediction for cold start recommender systems, where users have initially little to no information, and thus, their connections cannot be inferred. However, their connections can be inferred after they have consumed some items. Instead of waiting for feedback, we can use expanding graph models to infer the signal value and subsequently update the filter online. To address this setting, we first propose an online stochastic update for the filters via specific heuristic models. Then, we propose an adaptive stochastic approach that learns also from an ensemble of topological expansion models.


\subsection{Heuristic Stochastic Online Filtering}

We model the connectivity of node $v_t$ via random stochastic models. Specifically,  we use the existing topology $\bbA_{t-1}$ to fix the attachment probabilities $\bbp_t$ and weights $\bbw_t$ using a heuristic attachment rule. Given $\bba_{t}$ is a random vector, the environment reveals the statistical loss
%
%
\begin{align}
    \begin{split}
 l_t(\bbh,x_t)=\mathbb E[f_t(\bbh,x_t)]+r(\bbh)
    \end{split}
\end{align}
where the expectation concerns the stochastic attachment model. 
For $f_t(\bbh,x_t)$ being the squared loss, we have the mean squared error expression
\begin{align}\label{eq.stochLoss}
\begin{split}
    l_t(\bbh,x_{t})&=\E{\frac{1}{2}(\bba_{t}^{\top}\bbA_{x,t-1}\bbh-x_{t})^2}+\mu||\bbh||_2^2\\
    &= \frac{1}{2}((\bbw_t\circ\bbp_t)^{\top}\bbA_{x,t-1}\bbh-x_{t})^2\\
    &\quad+\frac{1}{2}(\bbA_{x,t-1}\bbh)^{\top}\bbSigma_{t}\bbA_{x,t-1}\bbh+\mu||\bbh||_2^2
\end{split}
\end{align}
%
where $\bbSigma_t$ is the attachment covariance matrix [cf. \eqref{eq.attachment_rule}]. The first term on the r.h.s. of \eqref{eq.stochLoss}, $s_t^2=\frac{1}{2}((\bbw_t\circ\bbp_t)^{\top}\bbA_{x,t-1}\bbh-x_{t})^2$ is the squared bias between the expected model output $(\bbw_t\circ\bbp_t)^{\top}\bbA_{x,t-1}\bbh$ and the true signal $x_{t}$. The second term $\frac{1}{2}(\bbA_{x,t-1}\bbh)^{\top}\bbSigma_{t}\bbA_{x,t-1}\bbh$ is the variance of the predicted output, and the third term penalizes a high $l_2$-norm of $\bbh$.\footnote{We could also consider adding a penalty parameter to the variance contribution if we want to tweak the bias-variance trade-off in the filter update.} The projected online gradient descent update of the filter parameters $\bbh(t)$ is
\begin{equation}\label{stochastic update}
    \bbh(t)=\underset{\ccalH}{\Pi}(\bbh(t-1)-\eta\nabla_h l_t(\bbh,x_t)|_{\bbh(t-1)})
\end{equation}
with gradient
\begin{align}\label{stochastic grad}
    \begin{split}
       \nabla_h l_t(\bbh,x_t) &= ((\bbw_t\circ\bbp_t)^{\top}\bbA_{x,t-1}\bbh-x_t)\bbA_{x,t-1}^{\top}\\
       &(\bbw_t\circ\bbp_t)+\bbA_{x,t-1}^{\top}\bbSigma_{t}\bbA_{x,t-1}\bbh+2\mu\bbh.
    \end{split}
\end{align}The stochastic loss [cf. \eqref{eq.stochLoss}] is differentiable and strongly convex in $\bbh$. Following Assumption \ref{assumption4}, the bias $s_t$, and the gradient \eqref{stochastic grad} are also upper-bounded, making the loss Lipschitz.
Algorithm \ref{stochastic algorithm} summarizes learning in this setting. The complexity of the online stochastic filter learning at time $t$ is of order $\ccalO(K(M_{t}+N_t))$. Check Appendix \ref{AppB} for further details. Note the dependency on $N_t$, the size of the graph at time $t$. Since we do not know the true attachment at the time of making the prediction, the stochastic attachment model assigns probabilities to each node, along with the weights. This leads to the dependance on $N_t$ while making the prediction [cf. \eqref{eq.stochLoss}]. This does not exist in the deterministic case, as we know $\bba_t$.

\smallskip
\noindent\textbf{Regret analysis}: To characterize the role of the stochastic topological model on the filter update, we compare the cumulative loss between the online stochastic update and the deterministic batch solution. This allows quantifying the performance gap by not knowing the attachment pattern. The regret reads as
\begin{equation}
\frac{1}{T}R_{s,T}(\bbh^{\star})=\frac{1}{T}\sum_{t=1}^Tl_t^s(\bbh^s(t-1),x_t)-l_t^d(\bbh^{\star},x_t)    
\end{equation}where $l_t^s(\cdot)$ denotes the stochastic loss and $l_t^d(\cdot)$ the deterministic loss. Similarly, $\bbh^d(t-1)$ and $\bbh^s(t-1)$ denote the online filter at time $t\!-\!1$ in the deterministic and stochastic settings, respectively. We claim the following.
\begin{theorem}\label{Prop 2}
\textit{At time $t$, let graph $\ccalG_{t-1}$ have $N_{t-1}$ nodes and $\bbh^s(t-1)$, $\bbh^d(t-1)$ be the filters learnt online in the stochastic and deterministic scenarios, respectively. Let the $n$th element of probability vector $\bbp_t$ be $[\bbp_t]_n$. Given Assumptions \ref{assumption1}-\ref{assumption4}, the Lipschitz constant $L_d$, and learning rate $\eta$, the normalized static regret for the stochastic setting is upper-bounded as}
\begin{align}\label{P2}
\small
    \begin{split}
       & \frac{1}{T}R_{s,T}(\bbh^{\star})\leq \frac{1}{T}\bigg(\sum_{t=1}^Tw_h^2Y^2(||\bbp_t||_2^2+M_{max})+\\&2Rw_hY\sqrt{||\bbp_t||_2^2\!+\!M_{max}}+w_h^2Y^2\bar{\sigma}_t^2\\&+ L_d||\bbh^s(t-1)-\bbh^d(t-1)||\bigg)+\frac{||\bbh^{\star}||_2^2}{2\eta}+\frac{\eta}{2}L_d^2T
    \end{split}
\end{align}
where $~\bar{\sigma}_t^2=\underset{n=1:N_{t-1}}{\text{max}}[\bbp_t]_n(1-[\bbp_t]_n)$ and $||\bbA_{x,t-1}\bbh||_2\leq Y$.
\end{theorem}
\par\noindent\textbf{Proof}: See Appendix \ref{app:sectionA}. \qed
\par \smallskip The regret bound in \eqref{P2} depends on the stochastic expanding model and the incoming data as follows:
\begin{itemize}
\item The sum of squared norms of the probability vectors corresponding to the attachment rule, via the terms $\sum_{t=1}^T||\bbp_t||_2^2$ and $\sum_{t=1}^T\sqrt{||\bbp_t||_2^2+M_{max}}$. This makes the choice of attachment probability $\bbp_t$ important as it influences the online learner. For example if $\bbp_t=\mathbf{1}_{N_{t-1}}$ for all $t$, the sum $\sum_{t=1}^T||\bbp_t||_2^2$ is of the order $T^2$, which means the regret bound diverges. Thus, the attachment rule should be selected such that $\sum_{t=1}^T||\bbp_t||_2^2$ is of order $\ccalO(T)$ or less. However, not all decaying attachment probabilities will reduce the bound reducing. It is necessary to have an inverse dependence on $N_t$, as is the case for the uniform distribution. This is for example the case of the uniformly at random attachment as we elaborate in Corollary \ref{Corollary2}.
\item The term $\frac{w_h^2Y^2}{T}\sum_{t=1}^T\bar{\sigma}_t^2$ is the sum of the maximum variance for an attachment rule $\bar{\sigma}_t^2$ over time. The maximum value of $\bar{\sigma}_t^2$ is $0.25$, attained for an attachment probability of $0.5$. For an attachment rule which has either high or low attachment probabilities per node, $\bar{\sigma}_t^2$ will be low, thus contributing less to the regret bound. This means a lower regret can result from stochastic attachment rules with a smaller uncertainty in attachment over the nodes.
\item The average distance between the stochastic and deterministic filters over the sequence $\frac{1}{T}\sum_{t=1}^T||\bbh^s(t-1)-\bbh^d(t-1)||_2$. If the filter trained with a stochastic attachment is further away from the filter updated with known attachment, the regret is higher. This can happen when the attachment rule cannot model the incoming node attachment and the filter prediction incurs a higher squared error. However, we can use this term to modify the filter update. One way to do this is to include a correction step to update the online filter after the true connection has been revealed. We will discuss this in Remark \ref{remark 2}.
\item The term $\frac{||\bbh^{\star}||_2^2}{2\eta T}+\frac{\eta}{2}L_d^2$ suggests similar factors which affect the deterministic regret will also affect the stochastic regret [cf. \eqref{regret}].
\end{itemize}
We now present how this regret bound reduces for the uniformly at random attachment.
\begin{corollary}\label{Corollary2}
Consider a uniformly at random attachment with $[\bbp_t]_n=\frac{1}{N_{t-1}}$. As the sequence length grows to infinity, i.e., $T\rightarrow\infty$, the regret upper bound becomes
\begin{align}\label{eq cor 2}
\small
    \begin{split}
        & \frac{1}{T}R_{s,T}(\bbh^{\star})\leq w_h^2M_{max}Y^2\!+\!Rw_hY(M_{max}\!+\!1)\\&+\frac{1}{T}\sum_{t=1}^T L_d||\bbh^s(t-1)-\bbh^d(t-1)||_2+\frac{||\bbh^{\star}||_2^2}{2\eta T}+\frac{\eta}{2}L_d^2
    \end{split}
\end{align}
\end{corollary}
\par\noindent\textbf{Proof~}: See Appendix \ref{AppD}.\qed\\
Corollary \ref{Corollary2} shows that the regret bound in \eqref{P2} can be improved upon with the right choice of attachment rules. Even though the attachment rule helps, there is a chance it fails to model the true attachment process, in which case the term $\frac{1}{T}\sum_{t=1}^T L_d||\bbh^s(t-1)-\bbh^d(t-1)||_2$ can diverge, ultimately, not making the learner not useful in the steady state.

\subsection{Adaptive Stochastic Online Filtering}
Oftentimes, a single attachment rule cannot describe the connectivity of the incoming nodes and an ensemble of stochastic rules is needed. This is seen in the regret bound in Theorem \ref{Prop 2}, which depends on the distance between the stochastic and deterministic online filters. This poses the additional challenge of how to combine these rules for the growing graph scenario. To tailor the combined rule to the online setting, we consider a linear combination of different attachment models and update the parameters as we do for the filter coefficients. Specifically, consider $M$ attachment rules parameterized by the probability vectors ${\{\bbp_{m,t}\}}_{m=1:M}$ and the corresponding weight vectors ${\{\bbw_{m,t}\}}_{m=1:M}$. Here, ${[\bbp_{m,t}]}_i$ denotes the probability of $v_{t}$ attaching to $v_i\in\ccalV_{t-1}$ under the $m$th rule and ${[\bbw_{m,t}]}_i$ the corresponding weight. Upon defining the dictionaries $\bbP_{t-1}=[\bbp_{1,t},\ldots,\bbp_{M,t}]\in\reals^{N_{t-1}\times M}$ and $\bbW_{t-1}=[\bbw_{1,t},\ldots,\bbw_{N,t}]\in\reals^{N_{t-1}\times M}$, we combine these models as
\begin{equation}\label{eq.composite_vectors}
    \barbp_t=\bbP_{t-1}\bbm~~\textnormal{and}~~ \barbw_t=\bbW_{t-1}\bbn
\end{equation}
where the combination parameters $\bbm$ and $\bbn$ belong to the probability simplex
\begin{equation}
    \ccal S^M =\{\bbalpha\in\reals^M,~\mathbf1_M^{\top}\bbalpha=1,~\bbalpha\succeq\mathbf0_M\}.
\end{equation}
with $\mathbf1_M$ being the vector of $M$ ones. For an existing node $v_i$, the $i$th row of $\bbP_t$ contains the corresponding rule-based probabilities. Equation \eqref{eq.composite_vectors} ensures that $\barbp_t$ represents a composite probability vector of attachment with $[\barbp_t]_i=\sum_{l=1}^Mm_l{[\bbp_{l,t}]}_i$ representing the probability of $v_t$ attaching to $v_i$. It also ensures that the weights in $\barbw_t$ lie in $[0,w_h]$. By representing the expanding graph model via the latent vectors $\bbm$ and $\bbn$, we can analyze them in lieu of the growing nature of the problem. This eases the setting as both $\barbp_t$ and $\barbw_t$ grow in dimensions with $t$ since learning these values directly becomes challenging.
\begin{algorithm}[!t]
	\caption{Adaptive stochastic online filtering (\textbf{Ada-OGF})}
	\begin{algorithmic}
 \STATE \textbf{Input:} Starting graph $\ccalG_0$, $\bbA_0$, $\bbx_0$, $\ccalT$
	\STATE \textbf{Initialization}: Pre-train $\bbh^s(0)$, Initialize $\bbm(0)=\mathbf1_M/M$, $\bbn(0)=\mathbf1_M/M$. Compute $\bbP_0$ and $\bbW_0$.
	\FOR{t=1:T}
	\STATE Obtain $v_t$, $\bar\bbp_t=\bbP_{t-1}\bbm(t-1)$, $\bar\bbw_t=\bbW_{t-1}\bbn(t-1)$
	\STATE\text{Prediction}:$(\bbW_{t-1}\bbn(t-1)\circ\bbP_{t-1}\bbm(t-1))^{\top}\bbA_{x,t-1}\bbh^s(t-1)$
	\STATE Reveal loss $l_t(\bbh,\bbm,\bbn,x_t)$
	\STATE Update $\bbh(t)$ following \eqref{h update ada}
	\STATE Update $\bbm(t)$ following \eqref{m update ada}
	\STATE Update $\bbn(t)$ following \eqref{n update ada}
	\STATE Reveal $\bba_t$, update $\bbA_t$, $\bbx_t$, $\bbP_{t}$, and $\bbW_t$
	\ENDFOR 
	\end{algorithmic} 
	\label{Algorithm Adaptive}
\end{algorithm}

\smallskip
\noindent\textbf{Online learner}: The instantaneous stochastic loss becomes
\begin{align}\label{ok}
\begin{split}
    l_t(\bbh,\bbm,\bbn,x_{t})&=\frac{1}{2}((\bbW_{t-1}\bbn\circ\bbP_{t-1}\bbm)^{\top}\bbA_{x,t-1}\bbh-x_{t})^2\\
    &+\frac{1}{2}(\bbA_{x,t-1}\bbh)^{\top}\barbSigma_{t}\bbA_{x,t-1}\bbh+\mu||\bbh||_2^2
\end{split}
\end{align}
where $\barbSigma_{t}=\diag((\bbW_{t-1}\bbn)^{\circ2}\circ(\bbP_{t-1}\bbm)\circ(\mathbf1_{N_{t-1}}-\bbP_{t-1}\bbm))$ is the covariance matrix of this adaptive method. We then proceed with an online alternating gradient descent over the filter parameters $\bbh$, the composite probability parameters $\bbm$, and the composite weight parameters $\bbn$ as
\begin{equation}\label{h update ada}
   \bbh(t) =\underset{\ccalH}{\Pi}(\bbh(t-1)-\eta\nabla_h l_t(\bbh,\bbm,\bbn,x_t)|_{\bbh(t-1)})\\
\end{equation}
\begin{equation}\label{m update ada}
    \bbm(t) =\underset{\ccalS^M}{\Pi}(\bbm(t-1)-\eta\nabla_m l_{t}(\bbh,\bbm,\bbn,x_t)|_{\bbm(t-1)})\\
\end{equation}
\begin{equation}\label{n update ada}
    \bbn(t) =\underset{\ccalS^M}{\Pi}(\bbn(t-1)-\eta\nabla_n l_{t}(\bbh,\bbm,\bbn,x_t)|_{\bbn(t-1)})
\end{equation}
where $\Pi_{\ccalS^M} (\cdot)$ is the projection operator onto the probability simplex $\ccalS^{M}$ and the gradient closed-form expressions are given in Appendix \ref{AppB}.
After the update, the environment reveals the true attachment $\bba_t$ and we update $\bbA_t$ and $\bbx_t$. We also update $\bbP_t$ based on the ensemble of attachment rules applied on the updated topology and the weight dictionary as $\bbW_t=[\bbW_{t-1};\bbe_t^{\top}]\in\reals^{N_t\times M}$ where $\bbe_t\in\reals^{M}$ contains independent positive random variables sampled uniformly between zero and the maximum possible edge weight $w_h$. Algorithm \ref{Algorithm Adaptive} highlights the adaptive stochastic online learning. The computational complexity at time $t$ for Ada-OGF is of order $\ccalO(K(M_0+N_t)+N_tM)$. See Appendix \ref{AppB} for more details.

The loss function in \eqref{ok} is jointly non-convex in $\bbh$, $\bbn$, and $\bbm$. It is marginally convex in $\bbn$ and $\bbh$ but not in $\bbm$ due to the nature of the covariance matrix. We can run multiple projected descent steps for each of the variables, but proving convergence is non-trivial. However, convergence to a local minimum of $l_t(\cdot)$ may not even be needed as we are in an online non-stationary setting where the arrival of another node leads to a new loss function. Thus, it is reasonable to take one or a few projected steps for each incoming node even without a full convergence guarantee.
\par\noindent\textbf{Regret analysis}: For the regret analysis of the stochastic adaptive online method, we claim the following.
\begin{corollary}\label{Corollary Ada}
   Given the hypothesis of Theorem \ref{Prop 2} and an adaptive stochastic online method over $M$ attachment rules with $\{\bbP_{t}\}$, the normalized static regret w.r.t. the deterministic batch learner is upper-bounded as
   \begin{align}\label{Regret adaptive}
     \small
    \begin{split}
        & \frac{1}{T}R_{s,T}(\bbh^{\star})\!\leq\! w_h^2Y^2\frac{1}{T}\sum_{t=1}^T\!\!\big(||\bbP_{t-1}||_F^2\!+\!M_{max}\big)\\&+R w_hY\frac{1}{T}\sum_{t=1}^T||\bbP_{t-1}||_2^2\!+Rw_hY(1+M_{max})+\!w_h^2Y^2\frac{1}{T}\sum_{t=1}^T\bar{P}_t\\&+\frac{1}{T}\sum_{t=1}^T L_d||\bbh^s(t-1)-\bbh^d(t-1)||_2+\frac{||\bbh^{\star}||_2^2}{2\eta T}+\frac{\eta}{2}L_d^2
    \end{split}
\end{align}where $\bar P_t=\underset{n=1:N_{t-1}}{\text{max}}||[\bbP_{t-1}]_{n,:}||_2$ and $M_{max}$ is the maximum number of edges formed by each incoming node.   
\end{corollary}
\par\noindent\textbf{Proof}: See Appendix \ref{AppC}.\qed
\par Compared to the single heuristic attachment model, the regret in \eqref{Regret adaptive} depends on the sum of $l_2$ norm squared of all the $M$ attachment rules. It also depends on $\bar{P}_t$, which is the maximum norm of the vector of probabilities for all rules for each node. 
\par\noindent The bound in \eqref{Regret adaptive} holds when selecting one attachment rule at each time, i.e.,  $||\bbm(t)||=1$ for all $t$. However, a smaller norm of $\bbm(t)$, corresponding to considering all rules leads to a lower regret bound, potentially improving the performance. Moreover, we expect the term concerning the distance between the stochastic and determinisitc filters to reduce due to the adaptive updates, thus, reducing the bound. We shall empirically corroborate this in Section \ref{Section Results}.
\begin{algorithm}[!t]
\caption{Prediction Correction Online Graph Filtering (\textbf{PC-OGF})}
\begin{algorithmic} 
\STATE \textbf{Input:} Graph $\ccalG_0$, $\bbA_0$, $\bbx_0$, $\ccalT = \ccalT=\{v_t,x_t,\bba_t\}_{t=1:T}$
\STATE \textbf{Initialize:} Pre-train $\bbh^s(0)$ over $\ccalG_0$ using $\bbA_0$, $\bbx_0$.
\FOR{$t=1:T$}
\STATE Obtain $v_t$ and $\bbp_t$, $\bbw_t$ following preset heuristics
\STATE Predict $\hat x_{t}=(\bbw_t\circ\bbp_t)^{\top}\bbA_{x,t-1}\bbh^s(t-1)$
\STATE Incur loss $l_t^s(\bbh,x_t)$ [cf. \eqref{eq.stochLoss}]
\STATE Update $\bbh^s(t)$ using \eqref{stochastic update}
\STATE Reveal $\bba_t$, update $\bbA_t$ and $\bbx_t$
\STATE Update $\bbh^s(t)$ using \eqref{online update}
\ENDFOR
\end{algorithmic}
\label{opc algorithm}
\end{algorithm}
\begin{remark}\label{remark 2}
\par\noindent\textit{Prediction Correction Online Graph Filtering} (\textbf{PC-OGF}): In the stochastic algorithms the bounds \eqref{P2} and \eqref{Regret adaptive} show that the regret is influenced by the difference between deterministic filters (that know the attachment) and the stochastic filters (that do not know the attachment) via the term $||\bbh^s(t-1)-\bbh^d(t-1)||_2$. One way to reduce the regret is to leverage the attachments after they are revealed and correct the learned stochastic filter coefficients via a deterministic update. 
This corresponds to using the prediction correction framework \cite{simonetto2017prediction}. The prediction step corresponds to performing the filter update based on the predicted output in the absence of connectivity information. The correction step performs an additional update on the prediction step by updating the filter for a loss function with the known attachment. The prediction and correction steps corresponds to one step of S-OGF and D-OGF, respectively. Algorithm \ref{opc algorithm} highlights this approach. The computational complexity of this at time $t$ is of order $\ccalO(K(M_{t}+N_t+M_{\textnormal{max}}))$, as it comprises one step of \textbf{S-OGF} followed by one of \textbf{D-OGF}.
\end{remark}

\section{Numerical Experiments}\label{Section Results}
We corroborate the proposed methods for regression tasks on both synthetic and real data-sets. We consider the following baselines and state-of-the-art alternatives.
\begin{enumerate}
    \item \textbf{D-OGF} [Alg. \ref{Online deterministic}]: This is the proposed online method for deterministic attachment. We search the filter order $K\in\{1,3,5,7,9\}$ and the learning rate $\eta$ and the regularization parameter $\mu$ from $[10^{-6},1]$.
    \item \textbf{S-OGF} [Alg. \ref{stochastic algorithm}]: This is the proposed online method using one stochastic attachment rule. We consider a uniformly at random attachment rule for $\bbp_t$. For $\bbw_t$, we use the same weight for each possible edge, which is the median of the edge weights in $\ccalG_{t-1}$. We obtain the regularization parameter $\mu$ and step-size $\eta$ via grid-search over $[10^{-5},10^{-1}]$.
    \item \textbf{Ada-OGF}: [Alg. \ref{Algorithm Adaptive}]. This is the proposed adaptive stochastic online method. We take $M=5$ with attachment rules based on the following node centrality metrics: $i)$ Degree centrality; $ii)$ Betweenness centrality \cite{newman2005measure}; $iii)$ Eigenvector centrality \cite{ruhnau2000eigenvector}; $iv)$ Pagerank; $v)$ Uniform.
    \item \textbf{PC-OGF} [Remark \ref{remark 2}]: This is the two-step update method. For the prediction step, we perform S-OGF with uniformly-at-random $\bbp_t$ and $\bbw_t$ as considered for S-OGF above. For the correction step, we perform one step of D-OGF. Both steps share the same learning rate $\eta\in[10^{-5},10^{-1}]$ and $\mu\in[10^{-5},10^{-1}]$.
    \item \textbf{Batch}: This is the filter designed by taking into account the whole node sequence, i.e., 
    \begin{equation}
        \bbh^{\star}=\underset{\bbh\in\reals^{K+1}}{\text{argmin}}\sum_{t=1}^T(\bba_{t}^{\top}\bbA_{x,t-1}\bbh-x_{t})^2+\mu||\bbh||_2^2
    \end{equation}which has a closed-form least-squares expression for $\mu>0\in[10^{-3},10]$.
    \item \textbf{Pre-trained}: This is a fixed filter trained over the existing graph $\ccal G_0$ and used for the expanding graphs. We train the filter over $80\%$ of the data over $\ccalG_0$. The regularization parameter is chosen over $[10^{-3},10]$.
    \item \textbf{OKL} \textit{Online Multi-Hop Kernel Learning} \cite{shen_online_2019}: We consider a Gaussian kernel with variance $\sigma^2\in\{0.1,1,10\}$. The number of trainable parameters is the same as that of the filters for a fair comparison.
    \item \textbf{OMHKL} \textit{Online Multi-Hop Kernel Learning} \cite{zong2021online}: This method considers multi-hop attachment patterns which are then fed into the random feature framework. We take the multi-hop length as the filter order. We consider one kernel for each hop with the same variance selected from $\sigma^2\in\{0.1,1,10\}$. We did not optimize over the combining coefficients for each multi-hop output. This is to keep the comparisons fair, as OMHKL has more parameters. Instead, we take the mean output, while updating the regression parameter for each multi-hop.
\end{enumerate}The hyper-parameters are chosen via a validation set. For each parameter, we perform a grid search over a specific range for each data-set, as indicated above for each approach. We use the same filter order as determined for \textbf{D-OGF} for the other online filters. We use the same filter order as determined for \textbf{D-OGF} for the other online filters. For all data sets, we divide the sequence of incoming nodes into a training and a test sequence. The first $80$ percent of the incoming node sequence are taken as the training nodes. The remaining $20$ percent are the test nodes. The nodes in the training sequence are used to tune the hyper-parameters, while the test set is used to evaluate the online method for the selected hyper-parameters.
\definecolor{Gray}{gray}{0.9}
\begin{table*}[t]
\centering
\caption{\label{NRMSE synthetic}Average NRMSE and standard deviation of all approaches for all data-sets.}
\begin{tabular}{l||c c||c c||c c|| c c|| c c}
\hline\hline
& \multicolumn{6}{c||}{\footnotesize{Synthetic Data}} & \multicolumn{4}{c}{\footnotesize{Real Data}}\\\hline
\footnotesize{Method} & \multicolumn{2}{c}{\footnotesize{Filter}} & \multicolumn{2}{c}{\footnotesize{WMean}} & \multicolumn{2}{c}{\footnotesize{Kernel}} & \multicolumn{2}{c}{\footnotesize{Movielens100K}} & \multicolumn{2}{c}{\footnotesize{ COVID}}\\
\hline
 & \footnotesize{NRMSE} &  \footnotesize{Sdev} & \footnotesize{NRMSE} & \footnotesize{Sdev} & \footnotesize{NRMSE} & \footnotesize{Sdev }& \footnotesize{NRMSE} & \footnotesize{Sdev} & \footnotesize{NRMSE} & \footnotesize{Sdev}\\ \hline
\rowcolor{Gray}
D-OGF (ours) & 0.02 & 0.003  & 0.02 & 0.005   & 0.25 & 0.04 & 0.26 & 0.01 & 0.21 & 0.02
\\
S-OGF (ours) & 0.18 & 0.02 & 0.26 & 0.06  & 0.28 & 0.07 & 0.28 & 0.007 & 0.31  & 0.02 
 \\
 \rowcolor{Gray}
Ada-OGF (ours) & 0.18 & 0.02  & 0.25 & 0.04  & 0.28 & 0.05 &  0.28 & 0.007  & 0.26 & 0.007
 \\
 PC-OGF (ours) & 0.18  & 0.02  & 0.22 & 0.02  & 0.23 & 0.04 &  0.27 & 0.01  & 0.26 & 0.003
 \\
  \rowcolor{Gray}
Batch & 0.04 & 0.007  & 0.09 & 0.04  & 1.3 & 0.29 & 6.7 & 0.1 & 0.17  & 0.03
 \\
\rowcolor{Gray}
 Pre-trained & 0.08 & 0.03  & 0.09 & 0.03  & 0.53 & 0.28 & 0.84 & 0.02 & 2.5 & 0.9 
 \\
  OKL & 0.17 & 0.01  & 0.23 & 0.02  & 0.25 & 0.04 &  0.27 & 0.01 & 0.25 & 0.02
 \\
\rowcolor{Gray}
 OMHKL & 0.17 & 0.01  & 0.32 & 0.1  & 0.34 & 0.09 & 0.27 & 0.01 & 0.25 & 0.02  \\
\hline\hline
\end{tabular}
\label{Table synthetic}
\end{table*}

\subsection{Experimental Setup}
We consider a synthetic setup based on a random expanding graph model; and two real data setups based on recommender systems and  COVID case predictions.
\par \smallskip
\noindent\textbf{Synthetic}: We start with a graph $\ccalG_0$ of $N_0=100$ nodes and an edge formation probability of $0.2$. The edge weights of $\bbA_0$ are sampled at random from the uniform distribution between zero and one. Each incoming node $v_t$ forms five uniformly at random edges with the existing graph $\ccalG_{t-1}$. Each newly-formed edge weight is the median of the edge weights in $\ccalG_0$. The existing graph signal $\bbx_0$ is band-limited w.r.t. the graph Laplacian, making it low-pass over $\ccalG_0$ with a bandwidth of three \cite{ortega_graph_2018}. We generate the true signal $x_t$ at the incoming $v_t$ in three ways to have three different types of data that fit the different methods.
\begin{enumerate}
\item \textit{Filter}: The true signal $x_t$ is generated using a pre-trained filter of order five on $\ccalG_0$. This setting is the closest to the proposed approach and is meant as a sanity check. It also helps us to investigate the differences between the deterministic and the stochastic attachments.
\item \textit{WMean}: $x_t$ is the weighted mean of the signals at the nodes $v_t$ attaches to. This is a neutral setting for all methods.
\item \textit{Kernel}: $x_t$ is obtained from a Gaussian kernel following \cite{shen_online_2019}. This prioritises kernel-based solutions and it is considered here as a controlled setting to compare our method in a non-prioritized setup.
\end{enumerate}
We average the performance of all methods over $10$ initial graphs $\ccalG_0$ and each having $T=1000$ incoming nodes with $800$ incoming nodes for training and $200$ for testing.

\smallskip
\noindent\textbf{Cold-start recommendation}: We consider the Movielens100K data-set that comprises $100,000$ ratings provided by $943$ users over $1152$ items \cite{harper2015movielens}. We build a $31$ nearest neighbour starting graph of $500$ random users and consider the remaining $443$ users as pure cold starters for the incoming sequence. We use the cosine similarity of the rating vectors to build the adjacency matrix of this graph. We use $50$ percent of the ratings of each new user $v_t$ to build $\bba_t$. 
We evaluated all methods over $10$ realizations of this setup, where, in each realization, we shuffle the order of incoming users. All methods perform online learning over 16875 and 6155 ratings in the training and test sets, respectively.

\par \smallskip
\noindent\textbf{COVID case prediction}: Here, we predict the number of COVID-19 infection cases for an uninfected city in an existing network of currently infected cities. We consider the data from \cite{dong2020interactive} that has daily case totals for $269$ cities and focus on a subset of $302$ days of this data-set as in \cite{giraldo2022reconstruction}. We randomly select $50$ cities and build a five nearest neighbour-directed graph $\ccalG_0$. The edge weight between cities $v_i$ and $v_j$ is $A_{ij}=\text{exp}(-\frac{||\bbt_i-\bbt_j||^2}{2\sigma^2})$, where $\bbt_i$ and $\bbt_j$ are the vector of  COVID cases from day one to $250$ for cities $v_i$ and $v_j$, respectively. We also use this interval to calculate the attachment vector $\bba_t$ for any incoming city node. We evaluate the performance on each of the days $255$, $260$, $265$, $270$, $275$, and $280$ and predict the  COVID case strength for each node city in the sequence. For each day, we carried out twenty realizations where we shuffle at random the order in which the cities are added to the starting graph.
 \par  We measure the performance via the root normalized mean square error NRMSE
\begin{equation}
\text{NRMSE}=\frac{\sqrt{\frac{1}{T}\sum_{t=1}^T(\hhatx_t-x_t)^2}}{\underset{t}{\textnormal{max}}~(x_t)-\underset{t}{\textnormal{min}}~(x_t)}.
\end{equation}
where $\hhatx_t$ and $x_t$ are the predicted and true signal at $v_t$, respectively. This measure gives a more realistic view of the performance since the incoming data does not follow a specific distribution and it is susceptible to outliers \cite{shcherbakov2013survey}. Additionally, we measure the normalized static regret (NReg) [cf. \eqref{regret basic}] for the online methods w.r.t. the Batch solution.

\subsection{Performance Comparison}

Table \ref{Table synthetic} comprises the NRMSEs and the standard deviations for all methods. We observe the following:
\par\noindent\textbf{Deterministic approaches}: D-OGF outperforms OKL and OMHKL across all the data-sets. The difference is more pronounced for the data generated using the \textit{Filter} and the \textit{WMean} method, as they are suited for filters, whereas for the \textit{Kernel} data, the difference is smaller. For the Movielens data-set, the difference is also small. We suspect this is because we train one filter across many graph signals (each graph signal corresponds to a different item) over the same user graph, whereas the kernel method ignores the graph signals. It is possible to improve the prediction accuracy by considering item-specific graphs as showcased in \cite{huang_rating_2018,das2022task}.
Next, we observe that D-OGF performs better than pre-trained throughout the experiments. This is because the online filters adapt to the incoming data stream, while the pre-trained does not. The only case we can expect a similar performance is where the incoming data is similar to the data over the existing graph. \par Concerning the batch solution, we find that the deterministic online learner outperforms Batch in all data-sets apart from the COVID data-set. This shows the limitations of batch-based solutions, i.e., an over-dependence on the observed training data, and also an inability to adapt to the sequence. For \textit{Filter} and \textit{WMean} data, the training and test set distributions are similar, so the difference between D-OGF and Batch can be attributed to the adaptive nature of D-OGF. In the other data-sets, the change in distribution is detrimental for the batch learner, particularly in the Movielens data.
\par\noindent\textbf{Stochastic approaches}: The S-OGF and Ada-OGF approaches have a similar performance for \textit{Filter}, \textit{Kernel} and Movielens data, with Ada-OGF performing better for \textit{WMean} and Covid data. This makes sense for the synthetic data as the constructed graphs expand following a uniformly at random attachment rule, the same rule used for \textbf{S-OGF}. The standard deviation is on the lower side for Ada-OGF. Since the existing signal $\bbx_0$ is band-limited, the signal values obtained via a filtering/ mean operation with a uniformly at random attachment will also be similar. However, Ada-OGF performs better for the  COVID data. This is because the incoming data in the  COVID data-set is quite different from the synthetic data. It does not have properties like smoothness and thus a uniformly at random attachment cannot help in predicting the number of cases. In such a setting, a more adaptive approach will help. For Movielens data, there is no difference between the two methods, possibly due to the high number of ratings.
\par\noindent\textbf{Deterministic vs stochastic}: The deterministic methods outperform the stochastic counterparts as expected. The gap is closer for the \textit{Kernel} and Movielens data. For Movielens this can be attributed to the subsequent filter updates that are done for different graph signals over a fixed graph. This can cause high prediction errors. The same holds also for the kernel method, as it is based on the same graph. Since the filter takes the signal into account, it might be affected more. 
\par In the Movielens, \textit{Kernel}, and  COVID data, the pre-trained filter does not update itself and is thus at a disadvantage, compared to the online methods. For  COVID data, the signal over the incoming node, i.e., the number of cases can be quite different from the signals over which the pre-trained filter is learnt, accounting for a higher error. Among the proposed online methods, the stochastic online methods perform poorly w.r.t pre-trained for \textit{Filter} and \textit{WMean} data. This is expected as the data distribution of the incoming data is similar to that over $\ccalG_0$ for these scenarios. 
\par The PC-OGF
method performs better than the stochastic methods for all data, showing the added value of correcting for the true attachment. It even outperforms D-OGF for \textit{Kernel} data.
\subsection{Analysis of Online Methods}
\begin{figure}[t]
\centering
\includegraphics[trim=0 190 0 200,clip,width=0.5\textwidth]{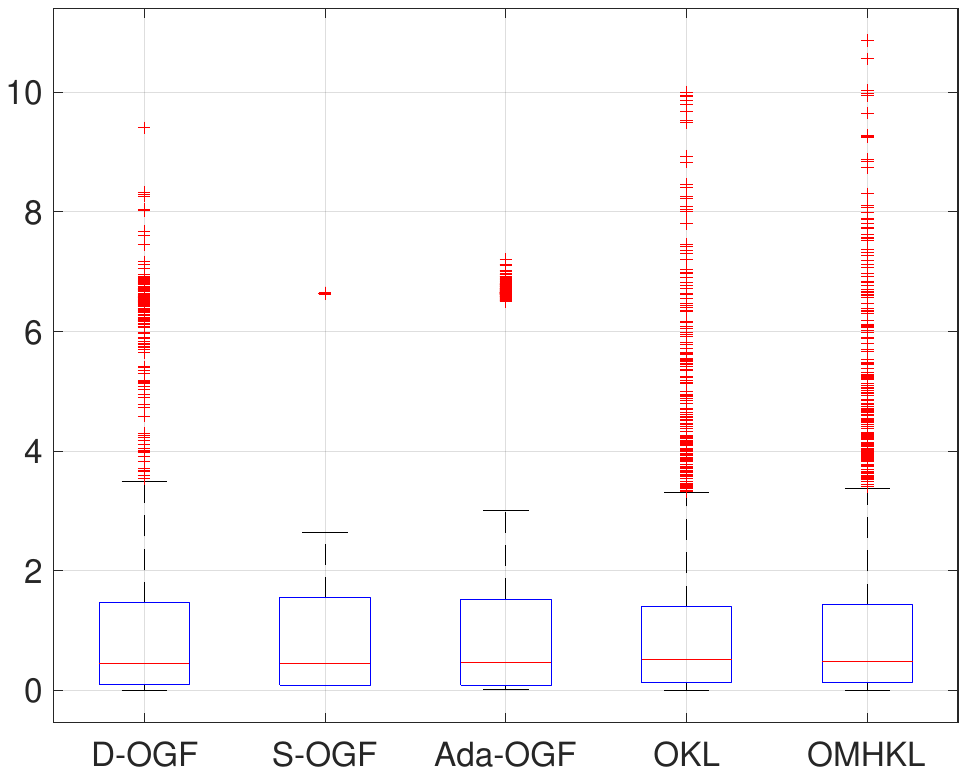}%
\caption{Box plot of the squared errors for each method in the Movielens data-set.}
\label{freq_plots_recsys} 
\end{figure}
\begin{figure}[t]
\centering
{\includegraphics[trim=30 210 50 200,clip,width=0.25\textwidth]{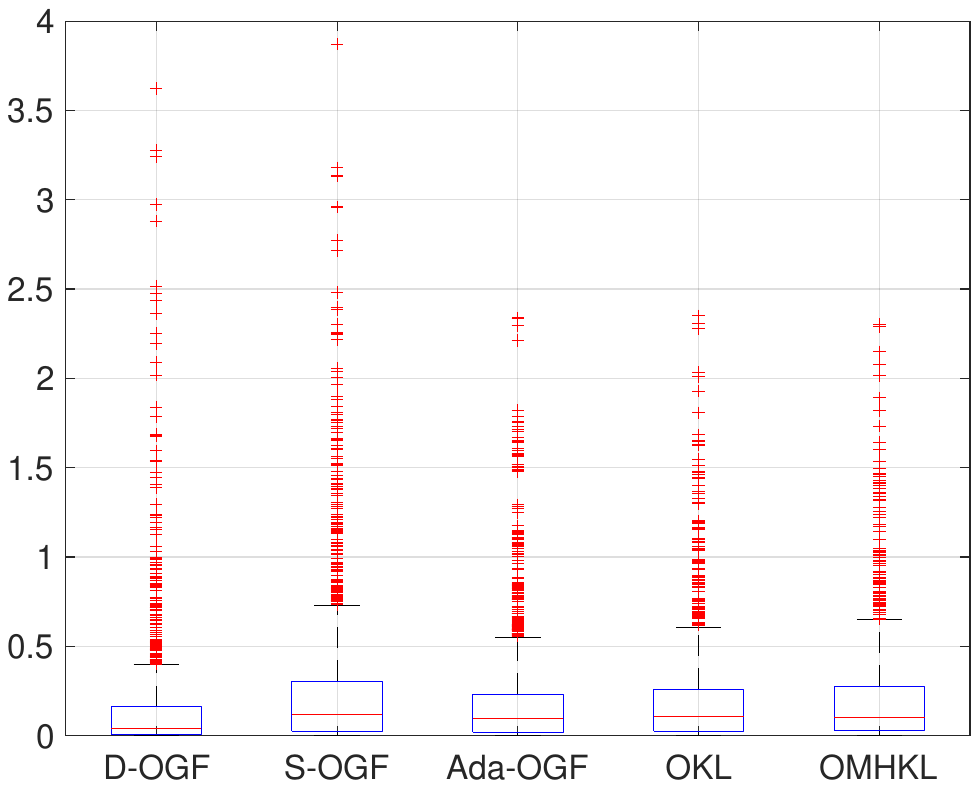}}%
{\includegraphics[trim=30 210 50 200,clip,width=0.25\textwidth]{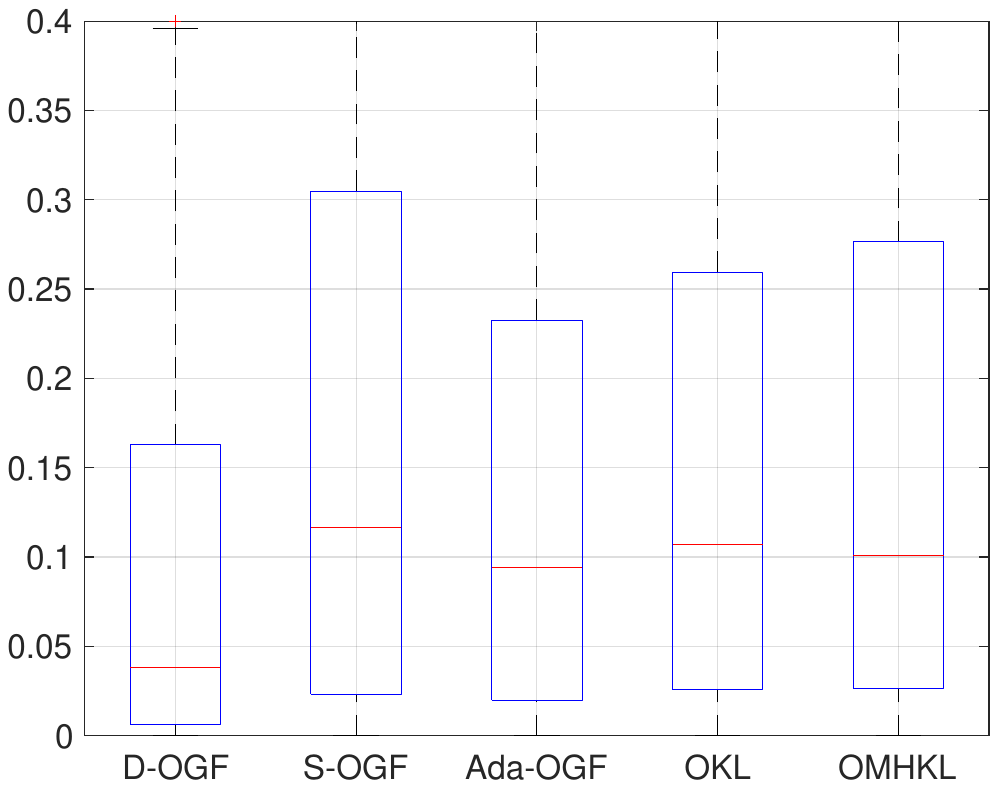}}%
\caption{(Left) Box plot of the squared errors across all data points over the six days. (Right) Box plot on the right zoomed in to highlight differences between all methods.}
\label{ COVID violin}
\end{figure}
We now investigate more in detail the online methods. 
\par\noindent\textbf{Outliers}: In Figs.~\ref{freq_plots_recsys} and \ref{ COVID violin} we show the violin plots of the squared errors over the test set for the Movielens and COVID data. The deterministic methods suffer more from higher outlier errors. This could be attributed to errors in estimating the attachment vectors. For Movielens data, we calculate this similarity over a subset of the items and for certain splits of the items this may lead to estimation errors for the similarity and thus also for the attachment vector. One reason why the stochastic methods are not prone to outliers could be the term in the loss functions [cf. \eqref{eq.stochLoss}, \eqref{ok}] that penalizes the prediction variance, ultimately, acting as a robust regularizer. Notably, for the Movielens data the errors in the stochastic online learners are fixed at certain levels. This is because the data-set has only five fixed values as ratings and because both S-OGF and Ada-OGF predict fixed values [cf. first term in \eqref{eq.stochLoss}]. For the Covid data, we calculated the number of outliers in the squared error. The outlier counts are $\textbf{D-OGF}=144$, $\textbf{S-OGF}=123$, $\textbf{Ada-OGF}=119$, $\textbf{OKL}=94$, $\textbf{OMHKL}=98$. This could also be due to the way the starting graph and the links of the incoming nodes are constructed. The figure on the right zooms in on the plot in the range between zero and $0.4$. The D-OGF has lower NRMSE, implying the presence of many samples with low squared error. The patterns for the other methods are similar.
\begin{figure*}[t!]
\centering
{\includegraphics[trim=20 210 50 200,clip,width=0.32\textwidth]{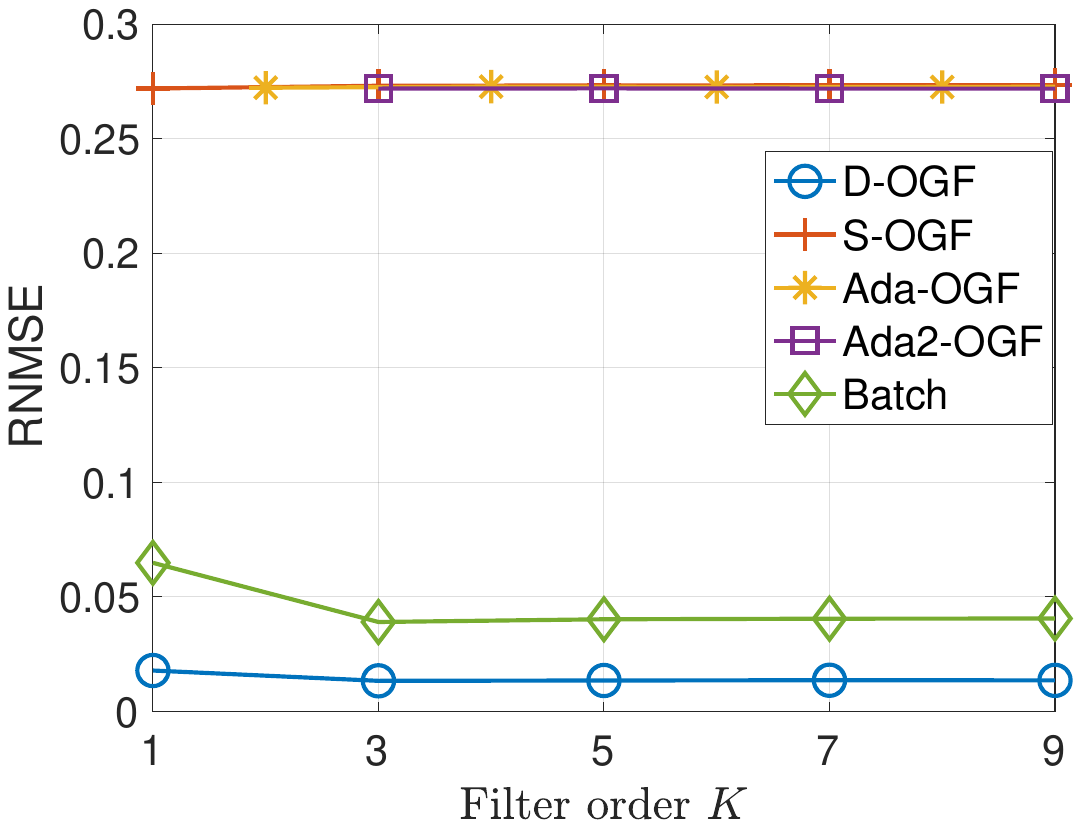}}%
{\includegraphics[trim=30 210 50 200,clip,width=0.32\textwidth]{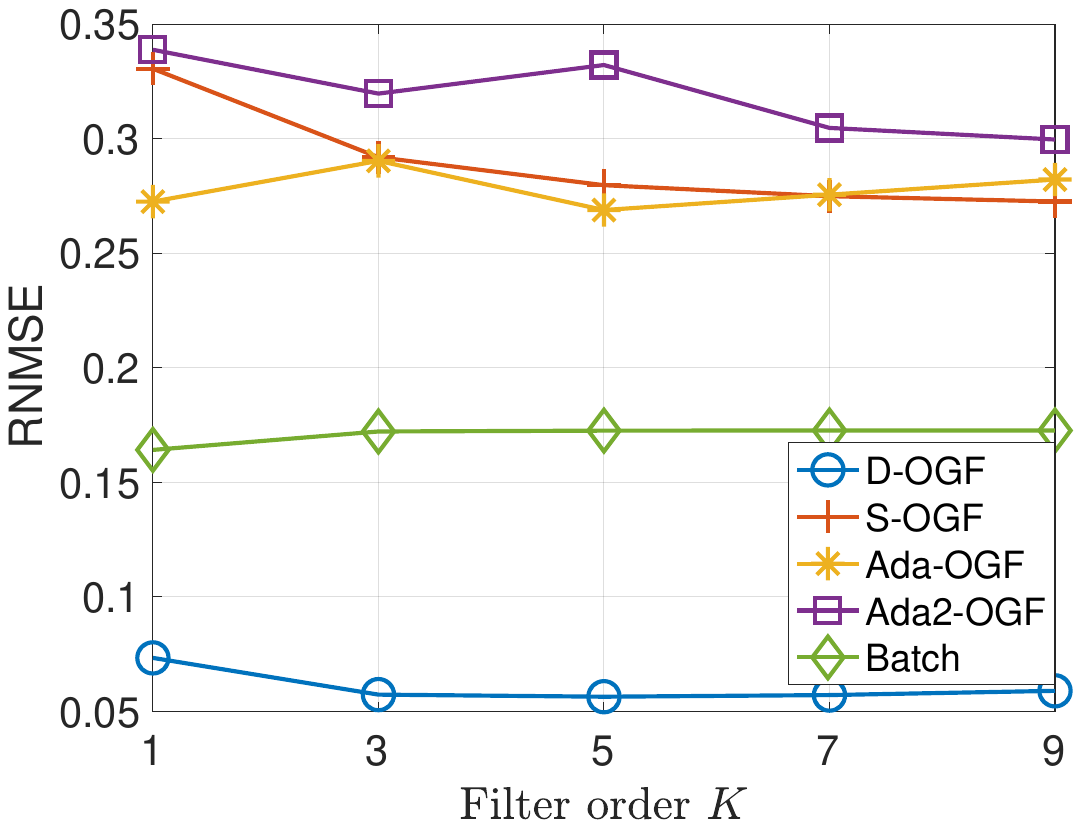}}%
{\includegraphics[trim=30 210 50 200,clip,width=0.32\textwidth]{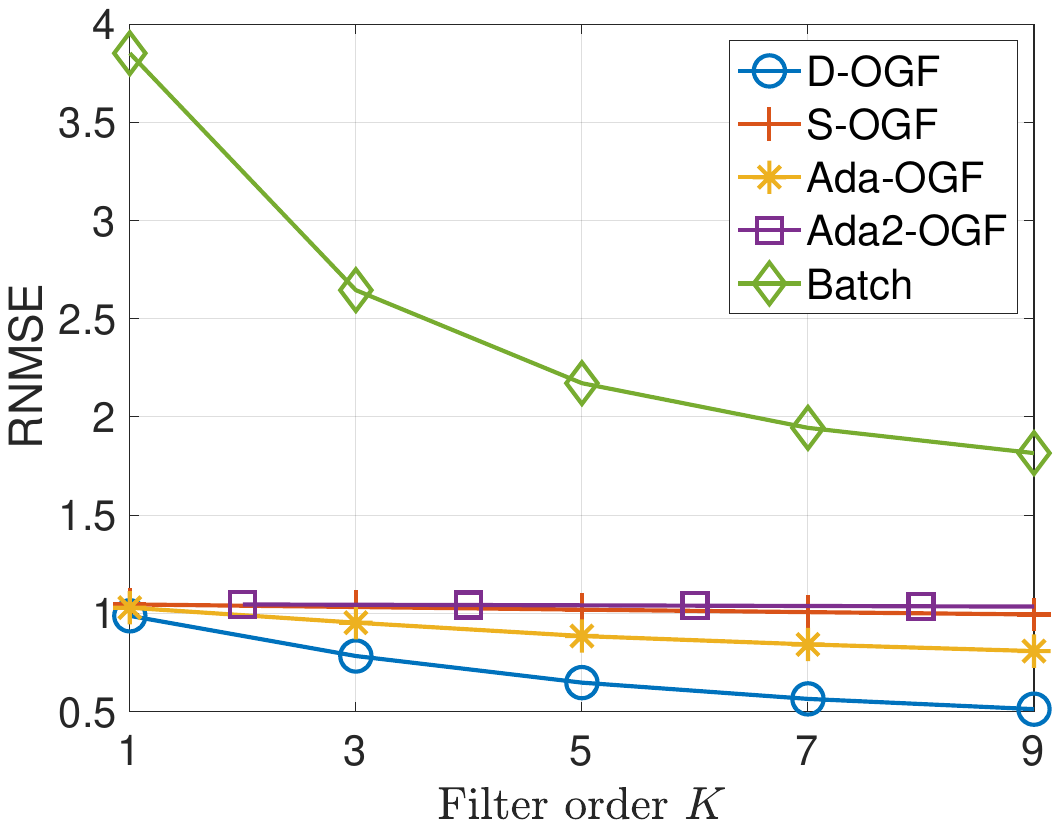}}%
\caption{RNMSE for different values of filter order $K$ for (left) \textit{Filter}, (centre) \textit{WMean}, and (right) \textit{Kernel} data, respectively.}
\label{Synthetic order dependence}
\end{figure*}
\par\noindent\textbf{Filter order, $\bbp$, $\bbw$}: Next, we investigate the role of the filter order as well as the impact of training both the attachment probabilities $\bbp_t$ and weights $\bbw_t$. Thus, we also want to compare with an alternative adaptive approach where we update only $\bbp$ while keeping $\bbw$ fixed to the true edge weights. We call this \textbf{Ada2-OGF}. We generate \emph{Filter} data, \emph{WMean} data, and \emph{Kernel} data with a variance of $10$. The filter orders evaluated are $K\in\{1,3,5,7,9\}$. Figure \ref{Synthetic order dependence} shows the variation of RNMSE of the filter approaches with filter order $K$. We see that \textbf{Ada2-OGF} performs worse than \textbf{Ada-OGF} apart from the \textit{Filter} data. This suggests that updating both $\bbp$ and $\bbw$ is beneficial than just updating $\bbp$. For the filter data, we see that all the three stochastic approaches perform the same. This is because we same stochastic rule, i.e., uniformly at random attachment for data generation. \textbf{S-OGF} uses the same, while \textbf{Ada-OGF} learns it.
\begin{figure*}[t!]
\centering
{\includegraphics[trim=20 210 50 200,clip,width=0.32\textwidth]{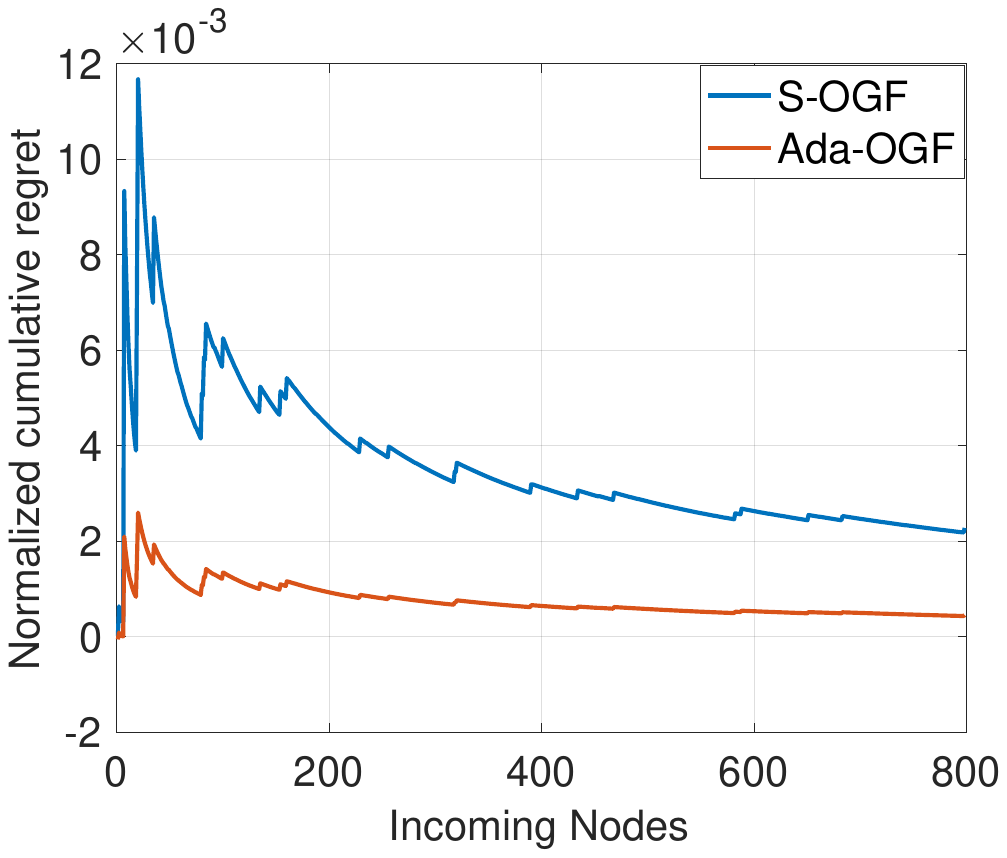}}%
{\includegraphics[trim=20 210 50 200,clip,width=0.32\textwidth]{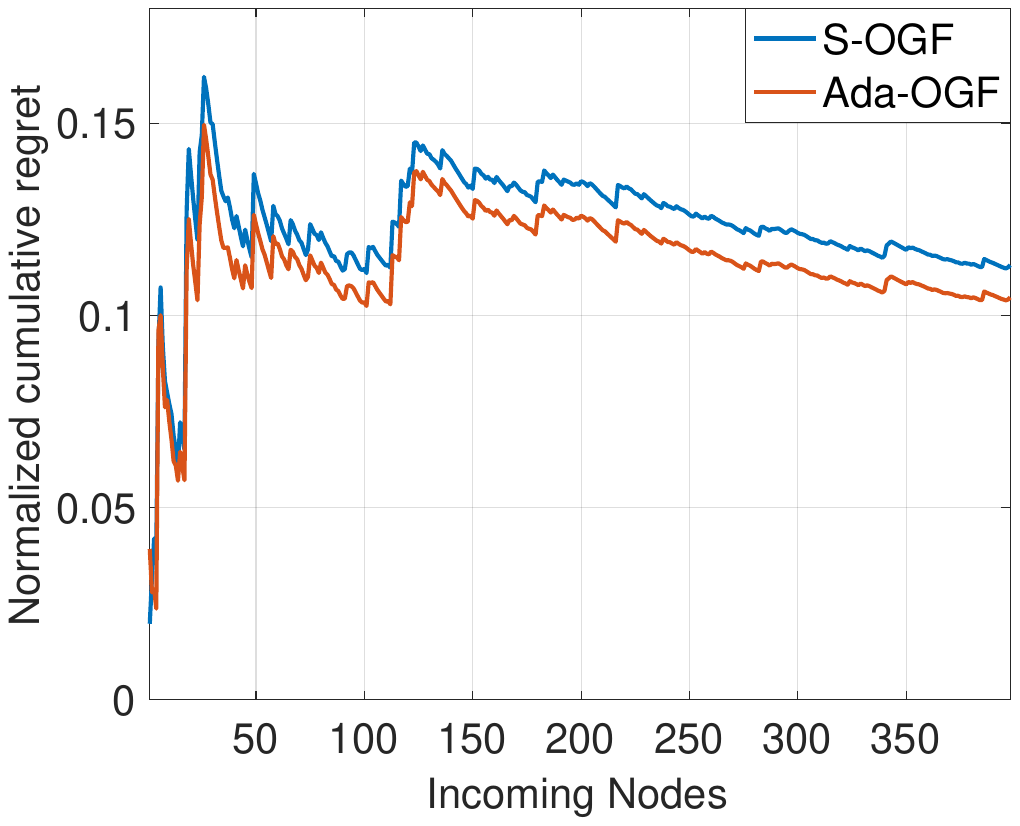}}%
{\includegraphics[trim=20 210 50 200,clip,width=0.32\textwidth]{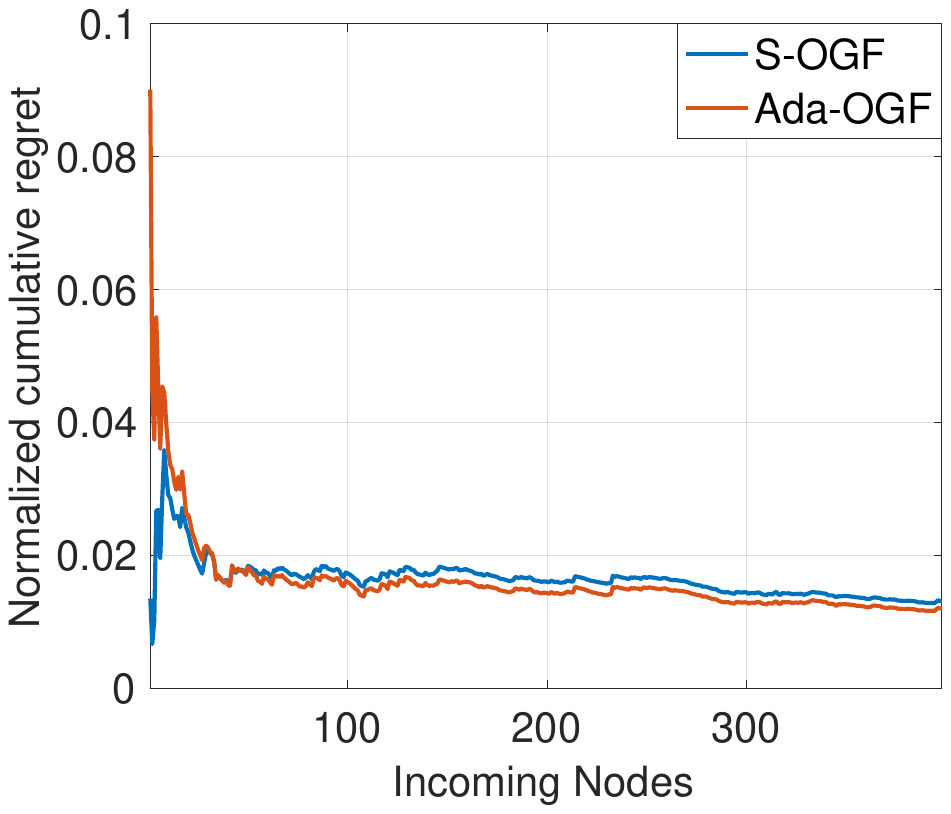}}%
\caption{Evolution of the normalized cumulative regret for S-OGF and Ada-OGF for the synthetic (left) \textit{Filter}, (center) \textit{WMean} and (right) \textit{Kernel} data for $T=800$, $T$ and $400$ incoming nodes, respectively.}
\label{cumulative regret stochastic}
\end{figure*}
\begin{figure*}[t!]
\centering
{\includegraphics[trim=20 210 50 200,clip,width=0.32\textwidth]{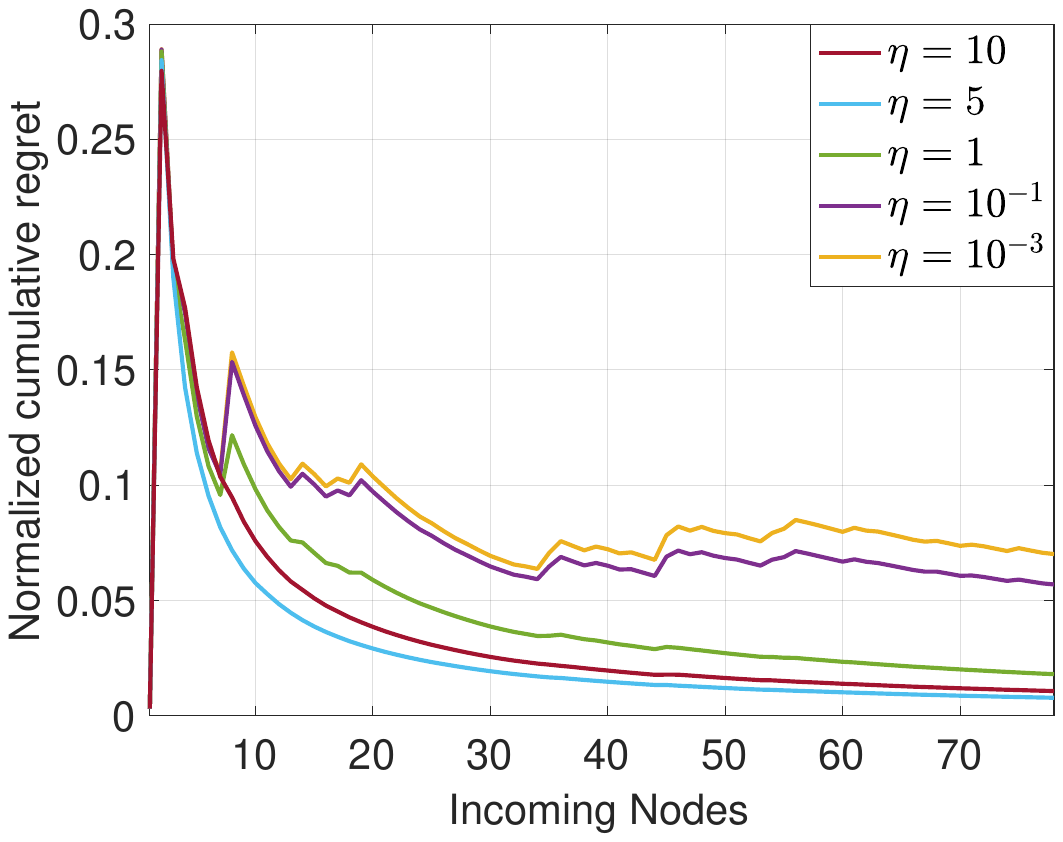}}%
{\includegraphics[trim=20 210 50 200,clip,width=0.32\textwidth]{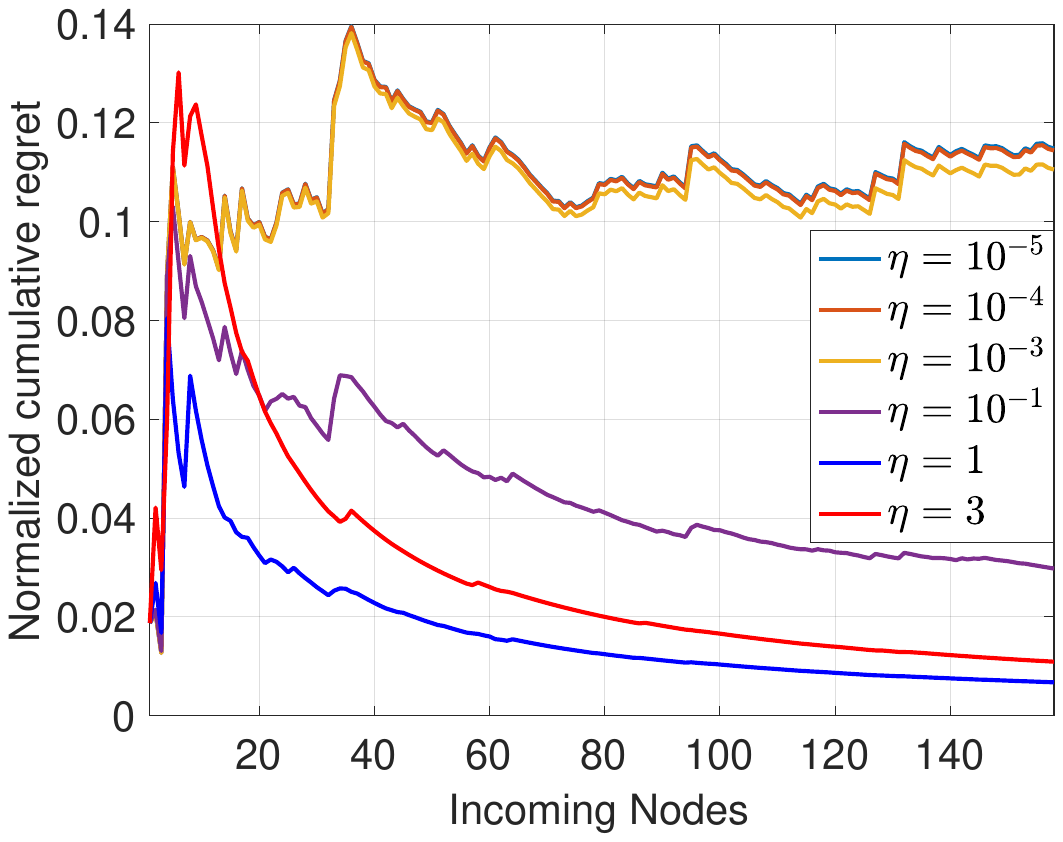}}%
{\includegraphics[trim=20 210 50 200,clip,width=0.32\textwidth]{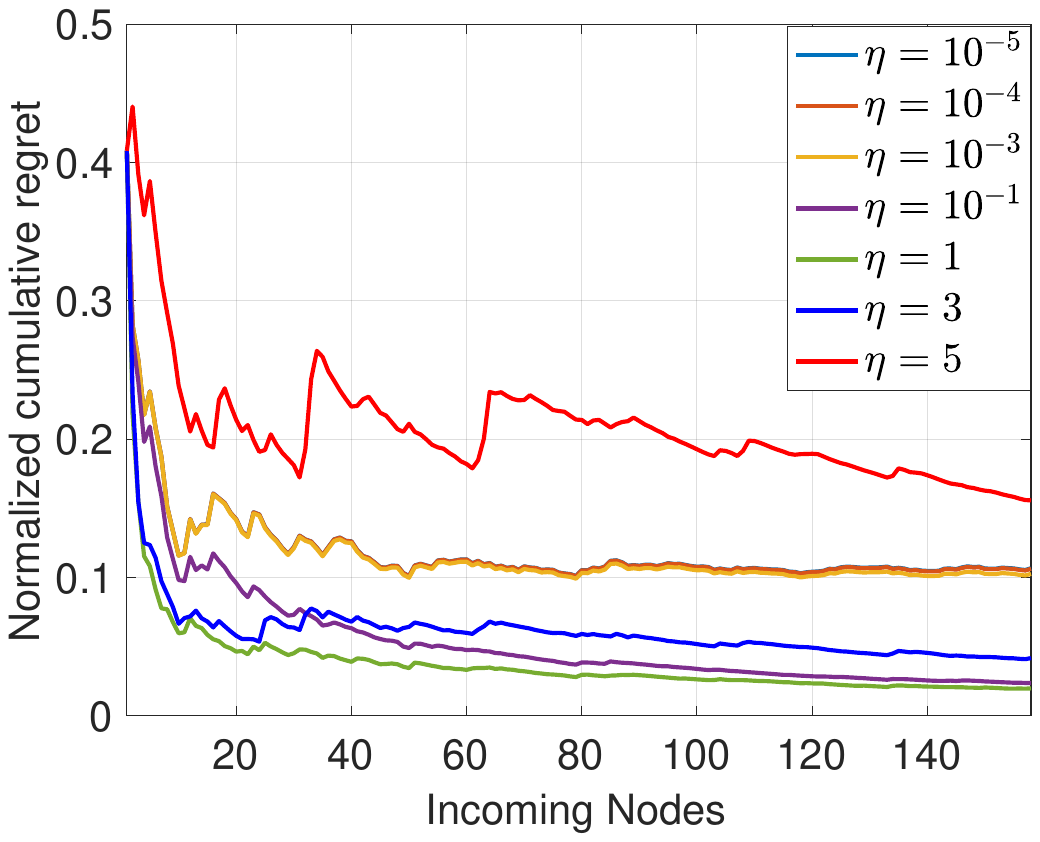}}%
\caption{Normalized cumulative regret evolution for different values of learning rate $\eta$ for (left) \textit{Filter}, (centre) \textit{WMean}, and (right) \textit{Kernel} data, respectively. The reference average error for the batch solution over the training set are $3\times10^{-5}$, $6.8\times10^{-4}$, and $1.1\times10^{-2}$, respectively.}
\label{effect of eta}
\end{figure*}
\par\noindent\textbf{Learning rate}: Figure \ref{effect of eta} shows the normalized cumulative regret at each time of $\textbf{D-OGF}$ w.r.t. the batch learner for different values of the learning rate $\eta$ for each synthetic-dataset. Increasing $\eta$  leads to a lower regret, but after one point, the regret increases. For the \textit{Kernel} data, for example, we see that the regret increases sharply between learning rate $\eta=3$ and $\eta=5$. This shows that $\eta$ indeed influences the online learner and its optimal value is in principle neither too high or too low. A higher value than the optimal misleads the online learner by focusing too much on the current sample. This can lead to high prediction errors for some samples, as seen in the spikes in the plots. A lower value learns about the incoming data-stream at a slower rate.
\par\noindent\textbf{Regret}: Figure \ref{cumulative regret stochastic} plots the normalized cumulative regret at each time for \textbf{S-OGF} and $\textbf{Ada-OGF}$ w.r.t. the batch solution for the \textit{Filter} (left), \textit{WMean} (center) and $\textit{Kernel}$ (right) data, respectively. In all three cases, the average cumulative regret converges, implying that the cumulative error or the gap with the batch solution does not diverge. This shows that the stochastic learners, despite not having access to the connectivity at the time of making a prediction, can learn from more incoming nodes. Second, \textbf{Ada-OGF} showcases a lower regret than \textbf{S-OGF}, showing that it can learn faster from the incoming nodes by trying to predict the attachment behaviour. This is in agreement with the regret bounds in Theorem \ref{Prop 2}, Corollaries \ref{Corollary2} and \ref{Corollary Ada}.
\par Finally, we investigate the normalized regret over the whole sequence for the online methods in Table \ref{Table Regret}. Since we evaluate this over the training set, we have positive values, which implies the batch solution has a lower cumulative error. However, having a positive regret during training can also lead to a lower NRMSE than the batch solution over the test set, as is the case for \textbf{D-OGF} [cf. Table $\textnormal{I}$]. This is because the batch filter is fixed and cannot perform as well as in the training set if the distribution of data in the test set is different. The lower regret for \textbf{D-OGF} compared to the stochastic approaches stems from the fact that the connectivity is known and because the Batch solution also has a similar loss function. The normalized regret for \textbf{PC-OGF} is lesser than that of the stochastic approaches, showing that incorporating the attachment can counter the effect of the gap between the stochastic and deterministic filter.
\begin{table}[t]
\centering
\caption{Normalized regret for the online methods for synthetic data.}
\begin{tabular}{l c c c}
\hline\hline
{\footnotesize{Method}} & {\footnotesize{Filter}} & {\footnotesize{WMean}} & {\footnotesize{Kernel}}\\
\hline
\rowcolor{Gray}
D-OGF & $1.6\times10^{-4}$ & 0.03 & 0.01 \\
S-OGF & $2.2\times10^{-3}$ & 0.84 & 0.08\\
 \rowcolor{Gray}
 Ada-OGF & $4.1\times10^{-3}$ & 0.82  & 0.11 \\
 PC-OGF & $1.9\times10^{-4}$ & 0.27 & 0.02\\
\hline\hline
\end{tabular}
\label{Table Regret}
\end{table}

\section{Conclusion}\label{Section Conclusion}
We proposed online filtering over graphs that grow sequentially over time. We adapted the formulation to the deterministic scenario where the connection of the incoming nodes is known and to a stochastic scenario where this connection is known up to a random model. We performed a simple projected online gradient descent for the online filter update and provided performance bounds in terms of the static regret. In the stochastic setting, the regret is a function of the rule-specific probabilities along with their variance. Numerical results for inference tasks over synthetic and real data show that graph filters trained online learning perform collectively better than kernel methods which do not utilize the data, pre-trained filters, and even a batch filter.
\par For future work, we will consider the scenario where the signal also varies over the existing graph, i.e., it has a spatio-temporal nature.
It is also possible to consider the scenario of joint topology and filter learning over the expanding graphs, where we estimate the true attachment of the incoming node instead of a stochastic model along with the filter used for making the inference. 
Finally, to account for the robustness of the online methods, one can also perform a weighted update, where the loss at a particular time is a weighted sum of the previous samples. The complexity of the stochastic approaches grow with the size of the graph. To tackle this, distributed filter updates can be a viable approach.
\appendices
\section{Proof of Theorem \ref{Prop 2}}
\label{app:sectionA}
The regret relative to the optimal filter $\bbh^{\star}$ is
\begin{align}
R_{s,T}(\bbh^{\star})=\sum_{t=1}^Tl_t^s(\bbh^s(t-1),x_t)-l_t^d(\bbh^{\star},x_t).
\end{align}By adding and subtracting the terms $\sum_{t=1}^Tl_t^d(\bbh^s(t-1),x_t)$ and $\sum_{t=1}^Tl_t^d(\bbh^d(t-1),x_t)$ we obtain
\begin{align}\label{eq2}
\begin{split}
&R_{s,T}(\bbh^{\star})=\sum_{t=1}^Tl_t^s(\bbh^s(t-1),x_t)-l_t^d(\bbh^s(t-1),x_t)
\\&+\sum_{t=1}^Tl_t^d(\bbh^s(t-1),x_t)-l_t^d(\bbh^d(t-1),x_t)
\\&+\sum_{t=1}^Tl_t(\bbh^d(t-1),x_t)-l_t^d(\bbh^{\star},x_t).
\end{split}
\end{align}where $l_t^d(\bbh^s(t-1),x_t)$ is the deterministic loss at time $t$ evaluated with the filter updated in the stochastic scenario. The regret in \eqref{eq2} comprises three sums over the $T$-length sequence, each of which contributes to the overall regret.
\par\noindent\textbf{The first term} in \eqref{eq2}, $\sum_{t=1}^Tl_{t}^s(\bbh^s(t-1),x_t)-l_t^d(\bbh^s(t-1),x_t)$ measures the difference in the stochastic and the deterministic loss for the filter updated online in the stochastic setting. We substitute
\begin{align}
l_t^s(\bbh^s(t-1),x_t)&=((\bbw_t\circ\bbp_t)^{\top}\bar{\bby}_t-x_t)^2+\bar{\bby}_t^{\top}\boldsymbol{\Sigma}_t\bar{\bby}_t\\&+\mu||\bbh^s(t-1)||_2^2
\end{align}where $\bar{\bby}_t=\bbA_{x,t-1}\bbh^s(t-1)$ and 
\begin{align}
l_t^d(\bbh^s(t-1),x_t)=(\bba_t^{\top}\bar{\bby}_t-x_t)^2+\mu||\bbh^s(t-1)||_2^2
\end{align}to get the difference at time $t$
\begin{align}
\begin{split}
&l_t^s(\bbh^s(t\!-\!1),x_t)\!-\!l_t^d(\bbh^s(t-1),x_t)\!=\!((\bbw_t\circ\bbp_t)^{\top}\bar{\bby}_t\!-\!x_t)^2\\&-(\bba_t^{\top}\bar{\bby}_t-x_t)^2+\bar{\bby}_t^{\top}\boldsymbol{\Sigma}_t\bar{\bby}_t.
\end{split}
\end{align}After some simplification, we get
\begin{align}\label{eq7}
\begin{split}
&l_t^s(\bbh^s(t-1),x_t)\!-\!l_t^d(\bbh^s(t-1),x_t)\!=\!((\bbw_t\circ\bbp_t\!-\!\bba_t)^{\top}\bar{\bby}_t)^2\\
&\!+\!2(\bbw_t\circ\bbp_t-\bba_t)^{\top}\bar{\bby}_t(\bba_t^{\top}\bar{\bby}_t-x_t)+\bar{\bby}_t^{\top}\boldsymbol{\Sigma}_t\bar{\bby}_t.
\end{split}
\end{align} The r.h.s. of equation \eqref{eq7} has three terms. For the first term we have
\begin{align}\label{00}
\begin{split}
  ((\bbw_t\circ\bbp_t-\bba_t)^{\top}\bar{\bby}_t)^2 & \leq ||\bbw_t\circ\bbp_t-\bba_t||_2^2||\bar{\bby}_t||_2^2 \\
  & \leq w_h^2(||\bbp_t||_2^2+M_{max})Y^2
\end{split}
\end{align}
 where the first inequality follows from the Cauchy-Schwartz inequality and the second inequality from Lemmas \ref{LM1} and Lemma \ref{lemma 3} in Appendix \ref{AppB}.
\par For the second term we have
\begin{align}\label{01}
\begin{split}
    & 2(\bbw_t\circ\bbp_t-\bba_t)^{\top}\bar{\bby}_t(\bba_t^{\top}\bar{\bby}_t-x_t)\\& \leq 2||\bbw_t\circ\bbp_t-\bba_t||_2||\bar{\bby}_t||_2||\bba_t^{\top}\bar{\bby}_t-x_t||_2\\
    & \leq 2Rw_hY\sqrt{||\bbp_t||_2^2+M_{max}}
    \end{split}
\end{align}where the first inequality follows from the Cauchy-Schwartz inequality and the second inequality from Assumption \ref{assumption4}, Lemmas \ref{lemma 3} and \ref{LM1}. For the third term, we have
\begin{align}\label{10}
\begin{split}
\bar{\bby}_t^{\top}\boldsymbol{\Sigma}_t\bar{\bby}_t&=\sum_{n=1}^{N_{t-1}}[\bar{\bby}_t]_n^2[\bbw_t]_n^2[\bbp_t]_n(1-[\bbp_t]_n)\\
&\leq w_h^2\bar{\sigma}_t^2Y^2
\end{split}
\end{align}where the inequality follows the definition of $\bar{\sigma}_t^2$ and Lemma \ref{lemma 3}. Adding \eqref{00}-\eqref{10} we can upper-bound \eqref{eq7} as
\begin{align}\label{Ineq1}
\begin{split}
&\sum_{t=1}^Tl_t^s(\bbh^s(t-1),x_t)\!-\!l_t^d(\bbh^s(t-1),x_t)
\\&\!\leq\! w_h^2Y^2({||\bbp_t||_2^2}\!+\!M_{max})\!+\!2Rw_hY\sqrt{||\bbp_t||_2^2\!+\!M_{max}}\\&+w_h^2\bar{\sigma}_t^2Y^2.
\end{split}
\end{align}
\par\noindent\textbf{The second term} in \eqref{eq2}, $\sum_{t=1}^Tl_{t}^d(\bbh^s(t-1),x_t)-l_t^d(\bbh^d(t-1),x_t)$ measures the sum of the differences in the deterministic loss between the deterministic and stochastic online filter. Since $l_t^d(\cdot,\cdot)$ is Lipschitz with constant $L_d$ from Lemma \ref{req}, we can write
\begin{align}
\begin{split}
&|l_t^d(\bbh^s(t\!-\!1),x_t)\!-\!l_t^d(\bbh^d(t\!-\!1),x_t)|
\\&\!\leq\! L_d||\bbh^s(t\!-\!1)\!-\!\bbh^d(t\!-\!1)||_2
\end{split}
\end{align}which implies $l_t(\bbh^s(t-1),x_t)-l_t(\bbh^d(t-1),x_t)\leq L_d||\bbh^s(t-1)-\bbh^d(t-1)||_2$. Summing over $t$, we have
\begin{align}\label{Ineq2}
\begin{split}
&\sum_{t=1}^Tl_t^d(\bbh^s(t-1),x_t)-l_t^d(\bbh^d(t-1),x_t)
\\&\leq L_d\sum_{t=1}^T||\bbh^s(t-1)-\bbh^d(t-1)||_2.
\end{split}
\end{align}
\par\noindent\textbf{The third term} in \eqref{eq2}, $\sum_{t=1}^Tl_t^d(\bbh^d(t-1),x_t)-l_t^d(\bbh^{\star},x_t)$ corresponds to the static regret in the deterministic case and has the upper bound 
\begin{align}\label{Ineq3}
\begin{split}
\sum_{t=1}^T\big(l_t(\bbh^d(t-1),x_t)-l_t^d(\bbh^{\star},x_t)\big)\leq \frac{||\bbh^{\star}||_2^2}{2\eta}+\frac{\eta}{2}L_d^2T
\end{split}
\end{align}as shown in Proposition 1. 
\par By summing equations \eqref{Ineq1}, \eqref{Ineq2}, and \eqref{Ineq3}, we obtain
\begin{align}\label{Ineq4}
    \begin{split}
        & R_{s,T}(\bbh^{\star})\leq \sum_{t=1}^Tw_h^2Y^2(||\bbp_t||_2^2+M_{max})+\\&2Rw_hY\sqrt{||\bbp_t||_2^2\!+\!M_{max}}+w_h^2\bar{\sigma}_t^2Y^2\\&+ L_d||\bbh^s(t-1)-\bbh^d(t-1)||\bigg)+\frac{||\bbh^{\star}||_2^2}{2\eta}+\frac{\eta}{2}L_d^2T
    \end{split}
\end{align}Finally, dividing both sides by $T$ and using Lemma \ref{lemma 3}, we complete the proof. \qed
\section{Proof of Corollary \ref{Corollary2}}
\label{AppD}
We substitute $\bbp_t=\frac{1}{N_{t-1}}\mathbf1_{N_{t-1}}$, in each term of the stochastic regret bound. For the first term we have
\begin{align}
\sum_{t=1}^Tw_h^2Y^2(||\bbp_t||_2^2+M_{max})\leq w_h^2Y^2\sum_{t=1}^T\frac{1}{N_{t-1}}+w_h^2M_{max}Y^2T
\end{align}Substituting $N_t=N_0+t-1$, we have
\begin{align}\label{tp}
& \sum_{t=1}^Tw_h^2Y^2(||\bbp_t||_2^2+M_{max})
\\& \leq w_h^2Y\sum_{t=1}^T\frac{1}{N_0+t-1}+w_h^2M_{max}Y^2T
\end{align}Next, we bound the summation $\sum_{t=1}^T\frac{1}{N_0+t-1}$ as \footnote{For a function $f(t)>0$, we have $\sum_{t=1}^Tf(t)dt\leq\int_{t=1}^Tf(t)dt.$}
\begin{align}
    \sum_{t=1}^T\frac{1}{N_0+t-1}\leq\int_{t=1}^T\frac{1}{t+N_0-1}dt
\end{align}which gives us
\begin{align}\label{log integral}
    \sum_{t=1}^T\frac{1}{N_0+t-1}\leq\log(T+N_0-1)-\log(N_0)
\end{align}Substituting \eqref{log integral} in \eqref{tp}, we have
\begin{align}
 &\sum_{t=1}^Tw_h^2Y^2(||\bbp_t||_2^2+M_{max})\leq \\& w_h^2Y^2(\log(T\!+\!N_0\!-1)\!-\!\log(N_0))\!+\!w_h^2M_{max}Y^2T 
\end{align}On dividing by $T$ and taking its limit to infinite, we have
\begin{align}\label{Cor 1 1}
 &\underset{T\rightarrow\infty}{\text{lim}}\frac{1}{T}\sum_{t=1}^Tw_h^2Y^2(||\bbp_t||_2^2+M_{max})\leq \!w_h^2M_{max}Y^2
\end{align}where the first term vanishes as $T$ grows faster than $\log(T)$.
\par For the second term we have
\begin{align}
&\sum_{t=1}^T2Rw_hY\sqrt{||\bbp_t||_2^2\!+\!M_{max}} \\& \leq 2Rw_hY\sum_{t=1}^T\frac{1}{2}(||\bbp_t||_2^2+M_{max}+1)\\&\leq Rw_hY(\sum_{t=1}^T||\bbp_t||_2^2+T(M_{max}+1))
\end{align}where we use the fact that the geometric mean is lesser than or equal to the arithmetic mean. Utilizing the fact that $\underset{T\rightarrow\infty}{\text{lim}}\frac{1}{T}\sum_{t=1}^T||\bbp_t||_2^2$ is equal to zero (as shown above in \eqref{Cor 1 1}), we have
\begin{align}\label{Cor 1 2}
&\underset{T\rightarrow\infty}{\text{lim}}\frac{1}{T}\sum_{t=1}^T2Rw_hY\sqrt{||\bbp_t||_2^2\!+\!M_{max}}\!\!\leq\!\! Rw_hY(M_{max}\!+\!1)
\end{align}For the third term $w_h^2Y^2\bar{\sigma}_t^2$, we have
\begin{align}
\sum_{t=1}^Tw_h^2\bar{\sigma}_t^2||\bar{\bby}_t||_2^2\!\leq\!w_h^2 Y \sum_{t=1}^T\frac{1}{N_{t-1}}(\!1\!\!-\!\!\frac{1}{N_{t-1}}\!)\!\leq\! w_h^2Y\sum_{t=1}^T\frac{1}{N_{t-1}} 
\end{align}which holds for the uniformly at random attachment rule. Given $\underset{T\rightarrow\infty}{\text{lim}}\frac{1}{T}\sum_{t=1}^T\frac{1}{N_{t-1}}=0$, the third term vanishes in the limit. Adding \eqref{Cor 1 1} and \eqref{Cor 1 2}, along with the other terms from the stochastic regret bound, we have the required bound for Corollary \ref{Corollary2}.\qed
\section{Proof of Corollary \ref{Corollary Ada}}
\label{AppC}
To prove this corollary, we start with the result of Proposition \ref{Prop 2}. For the first term, we have $||\bbp_t||_2^2=||\bbP_{t-1}\bbm(t-1)||_2^2$, which is upper bounded as $||\bbm(t-1)||_2^2||\bbP_{t-1}||_F^2$. The maximum value of $||\bbm(t-1)||_2^2$ is one for $\bbm(t-1)\in\ccalS_M$.
\par For the second term, we use the Arithmetic Mean-Geometric Mean inequality as in Corollary \ref{Corollary2} and use again $||\bbp_t||_2^2\leq||\bbP_{t-1}||_F^2$. 
Similarly, for the third term we have 
\begin{align}
    \begin{split}
        &\bar{\sigma}_t^2=\underset{n=1:N_{t-1}}{\text{max}}[\bbp_t]_n(1-[\bbp_t]_n)\\ 
        &=\underset{n=1:N_{t-1}}{\text{max}}[\bbP_{t-1}\bbm(t-1)]_n(1-[\bbP_{t-1}\bbm(t-1)]_n)\\
        &\leq \underset{n=1:N_{t-1}}{\text{max}}[\bbP_{t\!-\!1}\bbm(t-1)]_n
        \leq \bbarP_t||\bbm(t-1)||_2 \leq P_t
        \end{split}
\end{align}Substituting $\bar{\sigma}_t^2$ in the regret for the stochastic setting, we complete the proof. \qed
\section{Relevant Derivations}
\label{AppB}
\begin{lemma}\label{req}
\textit{Under Assumption~\ref{assumption1} and \ref{assumption3}, the loss function $l_t(\bbh,x_t)$ is Lipschitz in $\bbh$. That is, the $l_2$ norm of the gradient of the loss at time $t$ is upper-bounded as
\begin{align}
||\nabla_h l_t(\bbh,x_t)||_2\leq L_d
\end{align}}where $L_d=(RC+2\mu H)$ 
\end{lemma}

\textit{Proof}. We apply the Cauchy-Schawrtz inequality on the r.h.s. of \eqref{det gradient} and get
\begin{align}
\begin{split}
  ||\nabla_h l_t(\bbh,x_t)||_2&\!\leq\!|\bba_{t}^{\top}\bbA_{x,t-1}\bbh\!-\!x_{t}|||\bbA_{x,t-1}^{\top}\bba_{t}||_2\!\!+\!\!2\mu||\bbh||_2
  \\&\leq RC+2\mu H.
\end{split}
\end{align}where we use Assumption \ref{assumption1}.\qed\\
\begin{lemma}\label{LM1}
\textit{At time $t$, given the probability of attachment $\bbp_t$, the weight $\bbw_t$, and the attachment $\bba_t$. Let $N_{t-1}$ be the number of existing nodes. We have}
\begin{align}
||\bbw_t\circ\bbp_t-\bba_t||_2^2\leq w_h^2(||\bbp_t||_2^2+M_{max}).
\end{align}
\end{lemma}
\textit{Proof}. We can write the squared norm as
\begin{align}
||\bbw_t\!\circ\!\bbp_t\!-\!\bba_t||_2^2\!=\!\sum_{n=1}^{N_{t-1}}[\bbw_t]_n^2[\bbp_t]_n^2\!-\!2[\bbw_t]_n[\bbp_t]_n[\bba_t]_n\!+\![\bba_t]_n^2
\end{align}The second term in this summation is always negative, so we can write
\begin{align}
||\bbw_t\circ\bbp_t-\bba_t||_2^2\leq\sum_{n=1}^{N_{t-1}}[\bbw_t]_n^2[\bbp_t]_n^2+[\bba_t]_n^2
\end{align}Note that $\sum_{n=1}^{N_{t-1}}[\bba_t]_n^2\leq M_{max}w_h^2$ using Assumptions \ref{assumption1} and \ref{assumption2}; thus, we have
\begin{align}
||\bbw_t\circ\bbp_t-\bba_t||_2^2\leq w_h^2||\bbp_t||_2^2+M_{max}w_h^2
\end{align}\qed

\begin{lemma}\label{lemma 3}
    The term $||\bbA_{x,t-1}\bbh||_2$ is bounded in its $\ell_2$ norm for all $t$, i.e., $||\bbA_{x,t-1}\bbh||_2\leq Y$
\end{lemma}
\noindent\textbf{Proof.} From the expression of $\tilde{\bby}_t$ in \eqref{filter output}, we have
\begin{align}
   ||\tilde{\bby}_t||_2 & \leq ||\sum_{k=0}^Kh_{k}\bbA_{t-1}^{k}{\bbx}_t||_2+||\bba_{t}^{\top}\sum_{k=1}^Kh_{k}\bbA_{t-1}^{k-1}{\bbx}_t||_2 \\
   & \leq ||\sum_{k=0}^Kh_{k}\bbA_{t-1}^{k}{\bbx}_t||_2+||\bba_t||_2||\sum_{k=1}^Kh_{k}\bbA_{t-1}^{k-1}{\bbx}_t||_2
\end{align}The first term on the R.H.S. is bounded for a bounded $\bbh$ and $\bbA_{t-1}$. So is the second term, following Assumptions \ref{assumption1} and \ref{assumption2}. Thus both the output $\tilde{\bby_t}$ and $\sum_{k=1}^Kh_{k}\bbA_{t-1}^{k-1}{\bbx}_t$ are bounded. We denote the bound for $||\sum_{k=1}^Kh_{k}\bbA_{t-1}^{k-1}{\bbx}_t||_2$ as $Y$. \qed

\par\noindent\textbf{Gradients:}
Here we provide the expressions for the gradients w.r.t $\bbh$, $\bbm$, and $\bbn$ for the adaptive stochastic online learner.
\begin{align}\label{h grad adap}
    \begin{split}
       &\nabla_h l_t^s(\bbh,\bbm,\bbn,x_t)= ((\bbW_{t-1}\bbn\circ\bbP_{t-1}\bbm)^{\top}
       \\&\bbA_{x,t-1}\bbh-x_t)\bbA_{x,t-1}^{\top}(\bbw_t\circ\bbp_t)+\bbA_{x,t-1}^{\top}\barbSigma_{t}\bbA_{x,t-1}\bbh\\
       &+2\mu\bbh.
    \end{split}
\end{align}
The gradient w.r.t. $\bbm$ is
\begin{align}\label{m grad adap}
    \begin{split}
    &\nabla_m l_{t}(\bbh,\bbm,\bbn,x_t)\\
    &=\bbP_{t-1}^{\top}(\bbW_{t-1}\bbn\!\circ\!\bbA_{x,t-1}\bbh)(\bbW_{t-1}\bbn\!
    \circ\!\bbP_{t-1}\bbm)^{\top}
    \bbA_{x,t-1}\bbh
    \!-\!x_{t})\\&+\bbP_{t-1}^{\top}((\bbA_{x,t-1}\bbh)^{\circ2}\circ(\bbW_{t-1}\bbn)^{\circ2})\\
    &-2\bbP_{t-1}^{\top}(\bbP_{t-1}\bbm\circ(\bbA_{x,t-1}\bbh)^{\circ2}\circ(\bbW_{t-1}\bbn)^{\circ2}).
\end{split}
\end{align}
Finally, the gradient w.r.t $\bbn$ is
\begin{align}\label{n grad adap}
    \begin{split}
        &\nabla_n l_{t}(\bbh,\bbm,\bbn,x_t)\\
        &=\bbW_t^{\top}(\bbP_{t-1}\bbm\circ\bbA_{x,t-1}\bbh)((\bbW_{t-1}\bbn\circ\bbP_{t-1}\bbm)^{\top}\\&\bbA_{x,t-1}\bbh-x_{t})
    \end{split}
\end{align}\qed
\par
\par\noindent\textbf{Computational complexity:}
Here we provide the computational complexity of the online learners.
All methods rely on computing $\bbA_{x,t-1}=[\tilde{\bbx}_{t},\bbA_{t-1}\tilde{\bbx}_{t},\ldots,\bbA_{t-1}^{K-1}\tilde{\bbx}_{t}]$ at time $t$. We construct $\bbA_{x,t-1}$ with a complexity of order $\ccalO(M_{t-1}K)$ formed by shifting $\tilde{\bbx}_{t}$ $K\!-\!1$ times over $\ccalG_{t-1}$. At time $t-1$, we need not calculate $\bbA_{x,t}=[\tilde{\bbx}_{t+1},\bbA_{t}\tilde{\bbx}_{t+1},\ldots,\bbA_{t}^{K-1}\tilde{\bbx}_{t+1}]$ from scratch. From the structure of $\bbA_t$ [cf. \eqref{topology update}], we only need to calculate the diffusions over the incoming edges $K-1$ times, which amounts to an additional complexity of $\ccalO(M_{\textnormal{max}}K)$. The computational complexity of each online method is detailed next.

\textit{D-OGF}: The complexity of update \eqref{online update} is governed by the gradient, which depends on the output at time $\hat x_t=\bba_{t}^{\top}\bbA_{x,t-1}\bbh$.
It has a complexity of $\ccalO(M_{\textnormal{max}}K+M_{t-1}K)$, where the complexity for $\bbA_{x,t-1}\bbh$ is $M_{t-1}K$ and that for the diffusion over the newly formed edges is $\ccalO(M_{\textnormal{max}}K)$. 

\textit{S-OGF}: The only difference between S-OGF and D-OGF is that we use the expected attachment vector $(\bbw_t\circ\bbp_t)$ to calculate the output, which means the complexity incurred depends on $N_{t-1}$ for all $t$.


\textit{Ada-OGF}: The Ada-OGF, being a stochastic approach has already a complexity of $\ccalO(K(M_0+N_t))$. However, it has an extra complexity of $\ccalO(N_t(M))$ due to both the $\bbm$ and $\bbn$ update.

\bibliographystyle{IEEEtran}
\bibliography{ref}

\begin{thebibliography}{10}
\providecommand{\url}[1]{#1}
\csname url@samestyle\endcsname
\providecommand{\newblock}{\relax}
\providecommand{\bibinfo}[2]{#2}
\providecommand{\BIBentrySTDinterwordspacing}{\spaceskip=0pt\relax}
\providecommand{\BIBentryALTinterwordstretchfactor}{4}
\providecommand{\BIBentryALTinterwordspacing}{\spaceskip=\fontdimen2\font plus
\BIBentryALTinterwordstretchfactor\fontdimen3\font minus
  \fontdimen4\font\relax}
\providecommand{\BIBforeignlanguage}[2]{{%
\expandafter\ifx\csname l@#1\endcsname\relax
\typeout{** WARNING: IEEEtran.bst: No hyphenation pattern has been}%
\typeout{** loaded for the language `#1'. Using the pattern for}%
\typeout{** the default language instead.}%
\else
\language=\csname l@#1\endcsname
\fi
#2}}
\providecommand{\BIBdecl}{\relax}
\BIBdecl

\bibitem{das2022online}
B.~Das and E.~Isufi, ``Online filtering over expanding graphs,'' in \emph{IEEE
  Asilomar Conference on Signals, Systems and Computations, Pacific Grove,
  USA}, 2022.

\bibitem{dong2020graph}
X.~Dong, D.~Thanou, L.~Toni, M.~Bronstein, and P.~Frossard, ``Graph signal
  processing for machine learning: A review and new perspectives,'' \emph{IEEE
  Signal Processing Magazine}, vol.~37, no.~6, pp. 117--127, 2020.

\bibitem{berberidis2018adaptive}
D.~Berberidis, A.~N. Nikolakopoulos, and G.~B. Giannakis, ``Adaptive diffusions
  for scalable learning over graphs,'' \emph{IEEE Transactions on Signal
  Processing}, vol.~67, no.~5, pp. 1307--1321, 2018.

\bibitem{sandryhaila2013discrete}
A.~Sandryhaila and J.~M.~F. Moura, ``Discrete signal processing on graphs,''
  \emph{IEEE Transactions on Signal Processing}, vol.~61, no.~7, pp.
  1644--1656, 2013.

\bibitem{isufi2021accuracy}
E.~Isufi, M.~Pocchiari, and A.~Hanjalic, ``Accuracy-diversity trade-off in
  recommender systems via graph convolutions,'' \emph{Information Processing \&
  Management}, vol.~58, no.~2, p. 102459, 2021.

\bibitem{isufi2024graph}
E.~Isufi, F.~Gama, D.~I. Shuman, and S.~Segarra, ``Graph filters for signal
  processing and machine learning on graphs,'' \emph{IEEE Transactions on
  Signal Processing}, 2024.

\bibitem{sandryhaila2014discrete}
A.~Sandryhaila and J.~M.~F. Moura, ``Discrete signal processing on graphs:
  Frequency analysis,'' \emph{IEEE Transactions on Signal Processing}, vol.~62,
  no.~12, pp. 3042--3054, 2014.

\bibitem{romero2016kernel}
D.~Romero, M.~Ma, and G.~B. Giannakis, ``Kernel-based reconstruction of graph
  signals,'' \emph{IEEE Transactions on Signal Processing}, vol.~65, no.~3, pp.
  764--778, 2016.

\bibitem{barabasi_emergence_1999}
A.~L. Barabási and R.~Albert, ``\BIBforeignlanguage{en}{Emergence of {Scaling}
  in {Random} {Networks}},'' \emph{\BIBforeignlanguage{en}{Science}}, vol. 286,
  no. 5439, Oct. 1999, publisher: American Association for the Advancement of
  Science.

\bibitem{barabasi2016network}
A.-L. Barab{\'a}si \emph{et~al.}, \emph{Network science}.\hskip 1em plus 0.5em
  minus 0.4em\relax Cambridge university press, 2016.

\bibitem{wang2010graph}
Z.~Wang, Y.~Tan, and M.~Zhang, ``Graph-based recommendation on social
  networks,'' in \emph{2010 12th International Asia-Pacific Web
  Conference}.\hskip 1em plus 0.5em minus 0.4em\relax IEEE, 2010, pp. 116--122.

\bibitem{huang_rating_2018}
W.~Huang, A.~G. Marques, and A.~R. Ribeiro, ``Rating {Prediction} via {Graph}
  {Signal} {Processing},'' \emph{IEEE Transactions on Signal Processing},
  vol.~66, no.~19, pp. 5066--5081, Oct. 2018.

\bibitem{silva2019pure}
N.~Silva, D.~Carvalho, A.~C. Pereira, F.~Mour{\~a}o, and L.~Rocha, ``The pure
  cold-start problem: A deep study about how to conquer first-time users in
  recommendations domains,'' \emph{Information Systems}, vol.~80, pp. 1--12,
  2019.

\bibitem{ortega_graph_2018}
A.~Ortega, P.~Frossard, J.~Kovačević, J.~M.~F. Moura, and P.~Vandergheynst,
  ``Graph {Signal} {Processing}: {Overview}, {Challenges}, and
  {Applications},'' \emph{Proceedings of the IEEE}, vol. 106, no.~5, pp.
  808--828, May 2018.

\bibitem{erdos_evolution_1961}
P.~Erdos, ``On the evolution of random graphs,'' \emph{Bulletin of the
  Institute of International Statistics}, vol.~38, pp. 343--347, 1961.

\bibitem{orabona2019modern}
F.~Orabona, ``A modern introduction to online learning,'' \emph{arXiv preprint
  arXiv:1912.13213}, 2019.

\bibitem{hazan2016introduction}
E.~Hazan \emph{et~al.}, ``Introduction to online convex optimization,''
  \emph{Foundations and Trends{\textregistered} in Optimization}, vol.~2, no.
  3-4, pp. 157--325, 2016.

\bibitem{shen_online_2019}
Y.~Shen, G.~Leus, and G.~B. Giannakis, ``Online {Graph}-{Adaptive} {Learning}
  {With} {Scalability} and {Privacy},'' \emph{IEEE Transactions on Signal
  Processing}, vol.~67, no.~9, pp. 2471--2483, May 2019.

\bibitem{rahimi2007random}
A.~Rahimi and B.~Recht, ``Random features for large-scale kernel machines,''
  \emph{Advances in neural information processing systems}, vol.~20, 2007.

\bibitem{zong2021online}
Z.~Zong and Y.~Shen, ``Online multi-hop information based kernel learning over
  graphs,'' in \emph{ICASSP 2021-2021 IEEE International Conference on
  Acoustics, Speech and Signal Processing (ICASSP)}.\hskip 1em plus 0.5em minus
  0.4em\relax IEEE, 2021, pp. 2980--2984.

\bibitem{money2021online}
R.~Money, J.~Krishnan, and B.~Beferull-Lozano, ``Online non-linear topology
  identification from graph-connected time series,'' in \emph{2021 IEEE Data
  Science and Learning Workshop (DSLW)}.\hskip 1em plus 0.5em minus 0.4em\relax
  IEEE, 2021, pp. 1--6.

\bibitem{money2023sparse}
------, ``Sparse online learning with kernels using random features for
  estimating nonlinear dynamic graphs,'' \emph{IEEE Transactions on Signal
  Processing}, 2023.

\bibitem{shafipour2020online}
R.~Shafipour and G.~Mateos, ``Online topology inference from streaming
  stationary graph signals with partial connectivity information,''
  \emph{Algorithms}, vol.~13, no.~9, p. 228, 2020.

\bibitem{venkitaraman_recursive_2020}
A.~Venkitaraman, S.~Chatterjee, and B.~Wahlberg, ``Recursive {Prediction} of
  {Graph} {Signals} {With} {Incoming} {Nodes},'' in \emph{{International}
  {Conference} on {Acoustics}, {Speech} and {Signal} {Processing} ({ICASSP})},
  May 2020, pp. 5565--5569.

\bibitem{chen2014semi}
S.~Chen, F.~Cerda, P.~Rizzo, J.~Bielak, J.~H. Garrett, and
  J.~Kova{\v{c}}evi{\'c}, ``Semi-supervised multiresolution classification
  using adaptive graph filtering with application to indirect bridge structural
  health monitoring,'' \emph{IEEE Transactions on Signal Processing}, vol.~62,
  no.~11, pp. 2879--2893, 2014.

\bibitem{dornaika2017efficient}
F.~Dornaika, R.~Dahbi, A.~Bosaghzadeh, and Y.~Ruichek, ``Efficient dynamic
  graph construction for inductive semi-supervised learning,'' \emph{Neural
  Networks}, vol.~94, pp. 192--203, 2017.

\bibitem{jian2018toward}
L.~Jian, J.~Li, and H.~Liu, ``Toward online node classification on streaming
  networks,'' \emph{Data Mining and Knowledge Discovery}, vol.~32, no.~1, pp.
  231--257, 2018.

\bibitem{nassif2017graph}
R.~Nassif, C.~Richard, J.~Chen, and A.~H. Sayed, ``A graph diffusion lms
  strategy for adaptive graph signal processing,'' in \emph{2017 51st Asilomar
  Conference on Signals, Systems, and Computers}.\hskip 1em plus 0.5em minus
  0.4em\relax IEEE, 2017, pp. 1973--1976.

\bibitem{hua2020online}
F.~Hua, R.~Nassif, C.~Richard, H.~Wang, and A.~H. Sayed, ``Online distributed
  learning over graphs with multitask graph-filter models,'' \emph{IEEE
  Transactions on Signal and Information Processing over Networks}, vol.~6, pp.
  63--77, 2020.

\bibitem{nassif2018distributed}
R.~Nassif, C.~Richard, J.~Chen, and A.~H. Sayed, ``Distributed diffusion
  adaptation over graph signals,'' in \emph{2018 IEEE International Conference
  on Acoustics, Speech and Signal Processing (ICASSP)}.\hskip 1em plus 0.5em
  minus 0.4em\relax IEEE, 2018, pp. 4129--4133.

\bibitem{elias2020adaptive}
V.~R. Elias, V.~C. Gogineni, W.~A. Martins, and S.~Werner, ``Adaptive graph
  filters in reproducing kernel hilbert spaces: Design and performance
  analysis,'' \emph{IEEE Transactions on Signal and Information Processing over
  Networks}, vol.~7, pp. 62--74, 2020.

\bibitem{dasfiltering2020}
B.~Das and E.~Isufi, ``Graph filtering over expanding graphs,'' in \emph{IEEE
  Data Science Learning Workshop {Processing} ({DSLW})}, May 2022.

\bibitem{liu2018streaming}
X.~Liu, P.~C. Hsieh, N.~Duffield, R.~Chen, M.~Xie, and X.~Wen, ``Streaming
  network embedding through local actions,'' \emph{arXiv preprint
  arXiv:1811.05932}, 2018.

\bibitem{das2022learning}
B.~Das and E.~Isufi, ``Learning expanding graphs for signal interpolation,'' in
  \emph{ICASSP 2022-2022 IEEE International Conference on Acoustics, Speech and
  Signal Processing (ICASSP)}.\hskip 1em plus 0.5em minus 0.4em\relax IEEE,
  2022, pp. 5917--5921.

\bibitem{das2022task}
B.~Das, A.~Hanjalic, and E.~Isufi, ``Task-aware connectivity learning for
  incoming nodes over growing graphs,'' \emph{IEEE Transactions on Signal and
  Information Processing over Networks}, vol.~8, pp. 894--906, 2022.

\bibitem{schein2002methods}
A.~I. Schein, A.~Popescul, L.~H. Ungar, and D.~M. Pennock, ``Methods and
  metrics for cold-start recommendations,'' in \emph{Proceedings of the 25th
  annual international ACM SIGIR conference on Research and development in
  information retrieval}, 2002, pp. 253--260.

\bibitem{vlaski2023networked}
S.~Vlaski, S.~Kar, A.~H. Sayed, and J.~M. Moura, ``Networked signal and
  information processing: Learning by multiagent systems,'' \emph{IEEE Signal
  Processing Magazine}, vol.~40, no.~5, pp. 92--105, 2023.

\bibitem{shalev2012online}
S.~Shalev-Shwartz \emph{et~al.}, ``Online learning and online convex
  optimization,'' \emph{Foundations and Trends{\textregistered} in Machine
  Learning}, vol.~4, no.~2, pp. 107--194, 2012.

\bibitem{simonetto2017prediction}
A.~Simonetto and E.~Dall’Anese, ``Prediction-correction algorithms for
  time-varying constrained optimization,'' \emph{IEEE Transactions on Signal
  Processing}, vol.~65, no.~20, pp. 5481--5494, 2017.

\bibitem{newman2005measure}
M.~E. Newman, ``A measure of betweenness centrality based on random walks,''
  \emph{Social networks}, vol.~27, no.~1, pp. 39--54, 2005.

\bibitem{ruhnau2000eigenvector}
B.~Ruhnau, ``Eigenvector-centrality—a node-centrality?'' \emph{Social
  networks}, vol.~22, no.~4, pp. 357--365, 2000.

\bibitem{harper2015movielens}
F.~M. Harper and J.~A. Konstan, ``The movielens datasets: History and
  context,'' \emph{Acm transactions on interactive intelligent systems (tiis)},
  vol.~5, no.~4, pp. 1--19, 2015.

\bibitem{dong2020interactive}
E.~Dong, H.~Du, and L.~Gardner, ``An interactive web-based dashboard to track
  covid-19 in real time,'' \emph{The Lancet infectious diseases}, vol.~20,
  no.~5, pp. 533--534, 2020.

\bibitem{giraldo2022reconstruction}
J.~H. Giraldo, A.~Mahmood, B.~Garcia-Garcia, D.~Thanou, and T.~Bouwmans,
  ``Reconstruction of time-varying graph signals via sobolev smoothness,''
  \emph{IEEE Transactions on Signal and Information Processing over Networks},
  vol.~8, pp. 201--214, 2022.

\bibitem{shcherbakov2013survey}
M.~V. Shcherbakov, A.~Brebels, N.~L. Shcherbakova, A.~P. Tyukov, T.~A.
  Janovsky, V.~A. Kamaev \emph{et~al.}, ``A survey of forecast error
  measures,'' \emph{World applied sciences journal}, vol.~24, no.~24, pp.
  171--176, 2013.

\end{thebibliography}
\end{document}